\documentclass[letterpaper]{article} 

\usepackage{comment}  
\usepackage{paralist} 
\usepackage{booktabs}
\usepackage[table]{xcolor}
\usepackage{amssymb}
\usepackage{amsmath}
\usepackage{multirow}
\usepackage{enumitem}
\usepackage{tikz}
\usepackage{pgfplots}
\usepackage{subcaption}
\usepackage{xcolor}
\usepackage{colortbl}
\usepackage[table]{xcolor}
\usepackage{tabularx}
\usepackage{epigraph}
\definecolor{audiocolor}{HTML}{DDF0F7}   
\definecolor{neutralcolor}{HTML}{D6EAF8} 
\definecolor{biascolor}{HTML}{FADBD8}    
\definecolor{answercolor}{HTML}{D5F5E3}  
\definecolor{metacolor}{HTML}{F2F3F4}    
\usepackage{array}
\definecolor{tableheader}{RGB}{45, 62, 80}
\definecolor{tablerowalt}{RGB}{242, 245, 248}
\definecolor{headertext}{RGB}{255, 255, 255}
\definecolor{accentblue}{RGB}{31, 119, 180}
\definecolor{posstrong}{RGB}{198, 219, 239}
\definecolor{posmedium}{RGB}{107, 174, 214}
\definecolor{poslight}{RGB}{222, 235, 247}
\definecolor{neglight}{RGB}{252, 219, 199}
\definecolor{neutral}{RGB}{247, 247, 247}

\pgfplotsset{compat=1.18}
\usepackage{aaai2026}  
\usepackage{times}  
\usepackage{helvet}  
\usepackage{courier}  
\usepackage[hyphens]{url}  
\usepackage{graphicx} 
\usepackage{natbib}  
\usepackage{caption} 
\usepackage{algorithm}
\usepackage{algorithmic}

\usepackage{newfloat}
\usepackage{listings}
\DeclareCaptionStyle{ruled}{labelfont=normalfont,labelsep=colon,strut=off} 
\floatstyle{ruled}
\newfloat{listing}{tb}{lst}{}
\floatname{listing}{Listing}
\title{Seeing Is Not Deciding: \\ Can Multimodal LLMs Act as Effective CEOs?}
\author{
Yuyang Dai\textsuperscript{\rm 1}\qquad
Xueqing Peng\textsuperscript{\rm 3}\qquad
Yuxia Wang\textsuperscript{\rm 1}\\[0.35em]
Preslav Nakov\textsuperscript{\rm 2}\qquad
Zhuohan Xie\textsuperscript{\rm 2}
}
\affiliations{
\textsuperscript{\rm 1}INSAIT, Sofia University ``St. Kliment Ohridski'', Bulgaria\\[0.2em]
\textsuperscript{\rm 2}Mohamed bin Zayed University of Artificial Intelligence, Abu Dhabi, UAE\\[0.2em]
\textsuperscript{\rm 3}Yale University, New Haven, CT, USA\\[0.45em]
y9657422@gmail.com \quad xueqing.peng14@gmail.com \quad yuxia.wang@insait.ai\\[0.2em]
preslav.nakov@gmail.com \quad zhuohan.xie@mbzuai.ac.ae
}

\begin{document}

\nocopyright
\maketitle

\begin{abstract}
Large language models are increasingly applied as autonomous decision-making agents. However, in executive business decisions, existing benchmarks are limited to text-only settings. This makes it unclear whether models can perceive visual business evidence and effectively integrate it to improve decision quality.
We introduce \textsc{C-SuiteBench}, a controlled multimodal benchmark that includes five decision tasks under paired text-only and multimodal conditions across 50 scenarios.
We place nine frontier models in the role of a chief executive officer and evaluate their decision-making ability.
Multimodal inputs consistently improve evidence-centric reasoning, with the largest and most reliable gains appearing in risk forecasting and board-facing justification.
However, we uncover a \textbf{multimodal integration paradox}: adding visual business information \textit{degrades} constrained resource allocation for all nine models, even as visual grounding itself improves.
Ablation experiments reveal that this failure emerges from signal crowding, although each visual channel helps individually, their combination disrupts constraint satisfaction during decoding.
These findings demonstrate that visual perception and constrained action are separable bottlenecks in multimodal agents, and that indiscriminate visual augmentation can harm
high-stakes decision making, motivating selective grounding strategies for future executive AI systems.
\end{abstract}


\section{Introduction}

\begin{figure*}[t!]
\centering
\includegraphics[width=\linewidth]{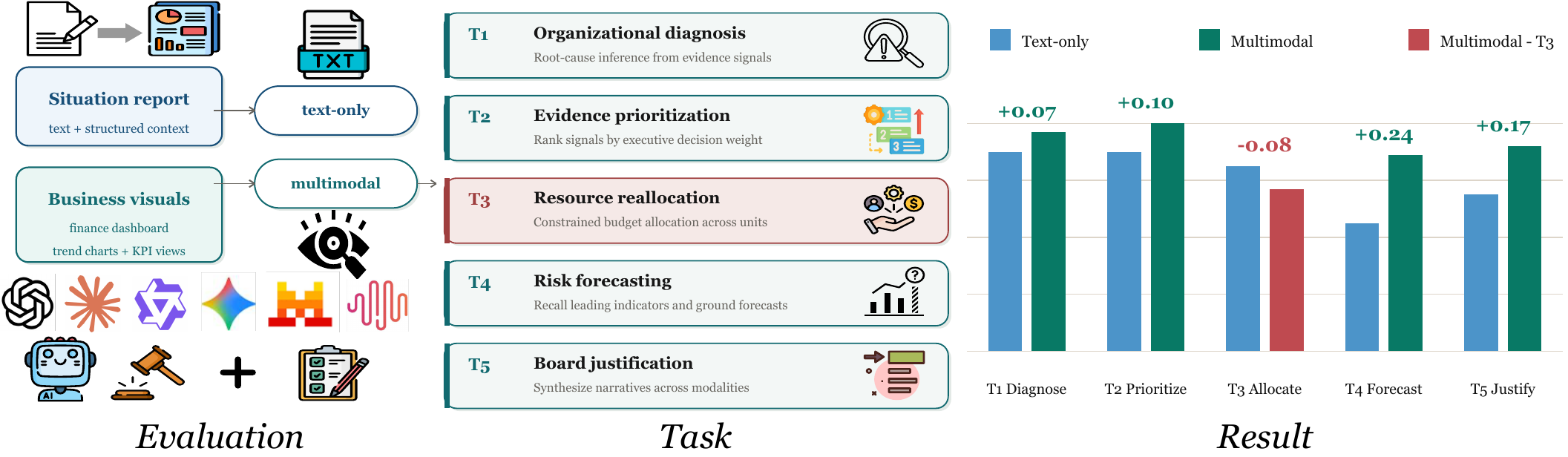}
\caption{Overview. \textbf{\textit{Left}} Each scenario pairs a textual situation report with rendered business visuals and is evaluated by task-specific automatic scoring under paired text-only and multimodal conditions.
\textbf{\textit{Middle}} The benchmark spans five executive tasks covering the full decision pipeline.
\textbf{\textit{Right}} Average task scores across all nine models.}
\label{fig:overview}
\end{figure*}

\epigraph{We don't see things as they are, we see them as we are.}{\textit{Ana\"{i}s Nin}}

Modern large language models (LLMs) exhibit increasingly sophisticated reasoning and planning capabilities, enabled by techniques such as chain-of-thought prompting~\citep{wei2022chain, zhang2022automatic}, ReAct~\citep{yao2022react}, and Tree of Thoughts~\citep{yao2023tree}. These advances have rapidly expanded to decision-making agents and led to strong performance in interactive decision-making environments~\citep{liu2024agentbench, yao2022webshop}.
Multimodal LLMs (MLLMs) have also achieved remarkable progress in visual understanding, including chart interpretation~\citep{masry2022chartqa, kahou2017figureqa, methani2020plotqa}, document parsing~\citep{mathew2021docvqa}, and expert-level visual question answering~\citep{yue2024mmmu, lu2024mathvista}. 
Together, MLLMs are becoming increasingly capable of supporting complex decisions grounded in both textual and visual evidence.
However, it remains unknown whether these advances can genuinely translate into practical, strategic decision support for Chief Executive Officers (CEOs). 
In this work, we focus on business-level decisions made by firms and CEOs.

Executive decisions (e.g.,\ whether to cut investment, expand production, or reassure investors) require integrating complex dimensions, such as financial dashboards, operational KPIs, growth trajectories, competitive dynamics, and long-term strategic objectives~\citep{hambrick1984upper, hambrick2007upper}. 
This makes the CEO-decision naturally multimodal. We aim to investigate whether advanced MLLMs with access to textual and visual evidence can effectively make executive decisions.

Existing evaluations treat reasoning and visual perception as separate capabilities. For general-purpose decision making,
agent benchmarks emphasize reasoning in predominantly text-based scenarios~\citep{liu2024agentbench, yao2022webshop, shridhar2020alfworld}, while multimodal benchmarks focus on chart reading and visual QA rather than strategic judgment~\citep{masry2022chartqa, kahou2017figureqa, methani2020plotqa, mathew2021docvqa, yue2024mmmu, lu2024mathvista}. 
For executive business decision making, financial benchmarks remain overwhelmingly text-centric, with nearly little visual business evidence~\citep{chen2021finqa, islam2023financebench, koncel2023bizbench, choi2025finagentbench, kundurthy2026bluefin, dai2026realfin}. This raises critical but unanswered questions: whether visual information improves executive decision-making, or the complexity of multimodal evidence integration introduces new failure modes that the text-only has not captured. 

To fill this gap, we introduce \textsc{C-SuiteBench}, a controlled benchmark for evaluating multimodal executive decision making. \textsc{C-SuiteBench} places MLLMs in the role of a chief executive officer reasoning over both textual reports and business dashboards, including financial performance charts, operational KPIs, and growth trend visualizations. The benchmark covers the full executive decision pipeline: organizational diagnosis, evidence prioritization, resource allocation, risk forecasting, and board justification, across \textit{\textbf{50 structured scenarios}} spanning \textit{\textbf{four levels of organizational complexity}}. Furthermore, To examine the contribution of visual information, we evaluate each scenario under paired \textit{text-only} and \textit{multimodal} conditions, keeping the underlying decision problem fixed. 
To sum up, our contributions are threefold. 
\begin{itemize}
    \item We introduce \textsc{C-SuiteBench}, the first controlled benchmark for evaluating whether visual business information improves executive decision making through paired text-only and multimodal conditions.
    \item We uncover a pattern of \textbf{task-asymmetric multimodal gains}, showing that visual evidence substantially benefits evidence-centric reasoning while degrading constrained resource allocation, revealing a \textbf{multimodal integration paradox} in modern MLLMs.
    \item We provide fine-grained analyses of how different categories of visual business evidence contribute to different stages of executive decision making, revealing that that each visual channel individually improves performance, yet combining all three channels degrades constrained resource allocation for every model tested while boosting risk forecasting across all nine models.
\end{itemize}

\section{Related Work}

\textbf{LLMs as Decision-Making Agents.}
Chain-of-thought prompting~\citep{wei2022chain,zhang2022automatic}, ReAct~\citep{yao2022react}, and Tree of Thoughts~\citep{yao2023tree} showed that LLMs can perform complex multi-step reasoning and sequential action. AgentBench~\citep{liu2024agentbench}, WebShop~\citep{yao2022webshop}, $\tau$-bench~\citep{yao2024tau}, and DSGBench~\citep{tang2025dsgbench} further demonstrate capable autonomous behavior across diverse task environments. These advances shift evaluation focus from whether models can reason to how they perform in realistic decision settings.

\noindent \textbf{Executive Decision-Making Benchmarks.}
CEO-Bench~\citep{chen2026ceo}, RetailBench~\citep{zhang2026retailbench}, and \citet{dai2026can} most directly evaluate strategic reasoning in organizational and business contexts. However, all existing executive benchmarks operate in text-only settings, leaving the role of visual business information entirely unexplored.

\noindent \textbf{Multimodal Business Visual Understanding.}
ChartQA~\citep{masry2022chartqa}, FigureQA~\citep{kahou2017figureqa}, DocVQA~\citep{mathew2021docvqa}, and MMMU~\citep{yue2024mmmu} assess whether models can correctly interpret visual artifacts. These benchmarks treat visual understanding as the primary focus, examining whether a model can interpret a chart rather than whether it can leverage the chart for downstream decision-making.

\noindent \textbf{Financial and Business Reasoning.}
FinQA~\citep{chen2021finqa}, FinanceBench~\citep{islam2023financebench}, BizBench~\citep{koncel2023bizbench}, and recent agentic extensions including BlueFin~\citep{kundurthy2026bluefin} and RealFin~\citep{dai2026realfin} cover a broad range of financial reasoning tasks. None provides paired text-only and multimodal evaluation over identical decision problems, making it impossible to isolate whether visual information genuinely improves strategic decisions (See Appendix~I).

\section{C-SuiteBench}

\begin{table*}[t]
\centering
\small
\renewcommand{\arraystretch}{1.1}
\setlength{\tabcolsep}{3pt}
\begin{tabular}{@{}clllc@{}}
\toprule
\rowcolor[HTML]{F5F4F0}
\textbf{Task} & \textbf{Decision Type} & \textbf{Key Visual Signal} & \textbf{Evaluation Dimensions} & \textbf{Avg.\ $\Delta$} \\
\midrule
\textsc{T1} & Root-cause identification   & Trend inflections, cross-metric divergence       & Diagnostic accuracy, evidence grounding          & \cellcolor[HTML]{EAF3DE}\textcolor[HTML]{3B6D11}{$+0.07$} \\
\textsc{T2} & Evidence ranking            & Signal salience, visual--textual conflict         & Ranking correlation, grounding fidelity          & \cellcolor[HTML]{EAF3DE}\textcolor[HTML]{3B6D11}{$+0.10$} \\
\textsc{T3} & Constrained reallocation    & Budget headroom, unit-level KPI trajectories      & Validity, reallocation accuracy, strategic fit   & \cellcolor[HTML]{FCEBEB}\textcolor[HTML]{A32D2D}{$-0.08$} \\
\textsc{T4} & Risk identification         & Leading-indicator trends, rate-of-change signals  & Risk recall, leading indicator recall, grounding & \cellcolor[HTML]{EAF3DE}\textcolor[HTML]{3B6D11}{$+0.24$} \\
\textsc{T5} & Stakeholder communication   & Cross-modal evidence synthesis                    & Narrative coherence, evidence integration        & \cellcolor[HTML]{EAF3DE}\textcolor[HTML]{3B6D11}{$+0.17$} \\
\bottomrule
\end{tabular}
\caption{The five executive tasks in \textsc{C-SuiteBench} and their average multimodal
uplift $\Delta = \text{score}_{\text{MM}} - \text{score}_{\text{text}}$ across all 9 models
($n = 50$ scenarios each).
T1 shows a positive directional trend ($p = 0.180$, not significant at model-level sign test);
T2, T4, T5 are statistically reliable ($p = 0.004$);
T3 is the only task with consistent degradation ($p = 0.004$).}
\label{tab:task_overview}
\end{table*}

\subsection{Overview}

To identify whether current MLLMs can translate the capability of information integration and reasoning across modalities into real-world executive decision making, we introduce \textsc{C-SuiteBench}.
This benchmark has two input settings: \textit{text-only} and \textit{multimodal}, while keeping the underlying decision problem fixed (Figure~\ref{fig:grand_overview}). 
This controlled design enables isolating the contribution of visual evidence in performance gains.

\textsc{C-SuiteBench} instantiates 50 formally structured scenarios spanning four organizational groups by difficulty (Section 3.2). 
Each scenario is evaluated across five executive tasks, yielding 250 scenario-task instances per input setting (500 in total for \textit{text-only} and \textit{multimodal}).
During the evaluation, the model to be assessed acts as a CEO and is presented with a structured situation report describing a company's financial position, operational status, and strategic context. Given the same decision problem, \textit{text-only} provides only the textual report, and \textit{multimodal} offers identical report but augmented with rendered business visuals, including financial performance charts, growth trend visualizations, and operational KPI dashboards.

We evaluated nine frontier MLLMs, with 2,250 paired text-only vs.\ multimodal task comparisons, showing a comprehensive measurement of multimodal uplift across tasks, difficulty levels, and model families.



\subsection{Scenario Design}
\label{sec:scenarios}

Each scenario in \textsc{C-SuiteBench} depicts a mid-sized enterprise at a strategic inflection point requiring executive judgment. Examples include revenue decline despite healthy operations, stalled growth with hidden unit-level risks, and conflicting departmental signals that require explicit resource trade-offs.
Scenarios were selected to cover organizational challenges where visual business evidence, financial performance trends, operational KPIs, and growth trajectories convey information that is difficult to recover from textual summaries alone, making them a natural testbed for multimodal executive evaluation.

To examine how the value of visual evidence varies with decision complexity, scenarios are organized into four difficulty families.
\begin{compactitem}
    \item \textbf{Easy}: consistent and unambiguous signals.
    \item \textbf{Fragile}: favorable surface indicators concealing underlying vulnerabilities.
    \item \textbf{Tension}: competing legitimate priorities requiring explicit trade-offs.
    \item \textbf{Adversarial}: intentionally misleading cues requiring models to distinguish relevant evidence from distractors.
\end{compactitem}
Within each scenario, text-only and multimodal conditions are constructed from the same underlying organizational state, ensuring that the two conditions differ only in the availability of visual business evidence rather than in the decision problem itself. Visuals are rendered directly from the scenario's underlying data. They expose quantitative patterns, trend inflections, rates of change, and cross-metric divergences, which are not explicitly stated in the textual report (full design details in Appendix~A.2).

\subsection{Task Design}
\label{sec:tasks}

Executive decision making is inherently a multi-stage process. Before taking action, executives must diagnose organizational problems, determine which evidence should guide decisions, translate those judgments into concrete actions, anticipate downstream risks, and communicate their rationale to stakeholders. Accordingly, \textsc{C-SuiteBench} decomposes executive reasoning into five complementary tasks, allowing multimodal gains and failures to be identified to specific stages of the decision-making pipeline.

\paragraph{T1: Organizational Diagnosis.} The model identifies the company's primary organizational challenge from the available evidence. As the first stage of executive reasoning, diagnosis lays the foundation for all subsequent decisions.

\paragraph{T2: Evidence Prioritization.} The model ranks candidate evidence items according to their decision relevance. Effective executives must determine not only what information is available, but also which signals deserve the greatest weight. 

\paragraph{T3: Resource Reallocation.} The model produces a concrete budget reallocation plan across organizational units under numerical constraints. Executive decisions ultimately require translating qualitative judgment into executable actions rather than remaining at the level of analysis alone. 

\paragraph{T4: Risk Forecasting.} The model identifies the most significant organizational risks over a long horizon and supports each prediction with an appropriate leading indicator. Strategic decision making requires anticipating future outcomes rather than reacting only to current conditions. 

\paragraph{T5: Board Justification.} The model produces a structured justification of its decisions for a board audience by integrating the evidence into a coherent executive narrative. Decisions are valuable only if they can be communicated and defended to stakeholders with competing priorities. 

\begin{table*}[t]
\centering
\renewcommand{\arraystretch}{1.1}
\setlength{\tabcolsep}{0pt}
\small
\begin{tabularx}{\textwidth}{@{}l
  *{3}{>{\centering\arraybackslash}X}
  *{3}{>{\centering\arraybackslash}X}
  *{3}{>{\centering\arraybackslash}X}
  *{3}{>{\centering\arraybackslash}X}
  *{3}{>{\centering\arraybackslash}X}@{}}
\toprule
\rowcolor[HTML]{F5F4F0}
 & \multicolumn{3}{c}{\textbf{T1\ \ Diagnose}}
 & \multicolumn{3}{c}{\textbf{T2\ \ Prioritize}}
 & \multicolumn{3}{c}{\textbf{T3\ \ Allocate}}
 & \multicolumn{3}{c}{\textbf{T4\ \ Forecast}}
 & \multicolumn{3}{c}{\textbf{T5\ \ Justify}} \\
\cmidrule(lr){2-4}\cmidrule(lr){5-7}\cmidrule(lr){8-10}
\cmidrule(lr){11-13}\cmidrule(lr){14-16}
\rowcolor[HTML]{F5F4F0}
\textbf{Model}
 & \small Text & \small MM & \small $\Delta$
 & \small Text & \small MM & \small $\Delta$
 & \small Text & \small MM & \small $\Delta$
 & \small Text & \small MM & \small $\Delta$
 & \small Text & \small MM & \small $\Delta$ \\
\midrule
GPT-5.4 mini
 & .648 & .674 & \cellcolor[HTML]{EAF3DE}\textcolor[HTML]{3B6D11}{$+$.03}
 & .735 & .817 & \cellcolor[HTML]{EAF3DE}\textcolor[HTML]{3B6D11}{$+$.08}
 & .720 & .645 & \cellcolor[HTML]{FCEBEB}\textcolor[HTML]{A32D2D}{$-$.08}
 & .567 & .787 & \cellcolor[HTML]{EAF3DE}\textcolor[HTML]{3B6D11}{$+$.22}
 & .613 & .806 & \cellcolor[HTML]{EAF3DE}\textcolor[HTML]{3B6D11}{$+$.19} \\
GPT-5.4 nano
 & .721 & .806 & \cellcolor[HTML]{EAF3DE}\textcolor[HTML]{3B6D11}{$+$.09}
 & .727 & .856 & \cellcolor[HTML]{EAF3DE}\textcolor[HTML]{3B6D11}{$+$.13}
 & .570 & .515 & \cellcolor[HTML]{FCEBEB}\textcolor[HTML]{A32D2D}{$-$.06}
 & .568 & .754 & \cellcolor[HTML]{EAF3DE}\textcolor[HTML]{3B6D11}{$+$.19}
 & .648 & .857 & \cellcolor[HTML]{EAF3DE}\textcolor[HTML]{3B6D11}{$+$.21} \\
Qwen2.5-VL 72B
 & .643 & .575 & \cellcolor[HTML]{FCEBEB}\textcolor[HTML]{A32D2D}{$-$.07}
 & .692 & .800 & \cellcolor[HTML]{EAF3DE}\textcolor[HTML]{3B6D11}{$+$.11}
 & .760 & .700 & \cellcolor[HTML]{FCEBEB}\textcolor[HTML]{A32D2D}{$-$.06}
 & .408 & .656 & \cellcolor[HTML]{EAF3DE}\textcolor[HTML]{3B6D11}{$+$.25}
 & .542 & .760 & \cellcolor[HTML]{EAF3DE}\textcolor[HTML]{3B6D11}{$+$.22} \\
Mistral Small
 & .571 & .707 & \cellcolor[HTML]{EAF3DE}\textcolor[HTML]{3B6D11}{$+$.14}
 & .709 & .808 & \cellcolor[HTML]{EAF3DE}\textcolor[HTML]{3B6D11}{$+$.10}
 & .580 & .520 & \cellcolor[HTML]{FCEBEB}\textcolor[HTML]{A32D2D}{$-$.06}
 & .448 & .716 & \cellcolor[HTML]{EAF3DE}\textcolor[HTML]{3B6D11}{$+$.27}
 & .580 & .795 & \cellcolor[HTML]{EAF3DE}\textcolor[HTML]{3B6D11}{$+$.22} \\
Claude Opus 4.8
 & .714 & .812 & \cellcolor[HTML]{EAF3DE}\textcolor[HTML]{3B6D11}{$+$.10}
 & .735 & .761 & \cellcolor[HTML]{EAF3DE}\textcolor[HTML]{3B6D11}{$+$.03}
 & .685 & .620 & \cellcolor[HTML]{FCEBEB}\textcolor[HTML]{A32D2D}{$-$.07}
 & .420 & .432 & \cellcolor[HTML]{EAF3DE}\textcolor[HTML]{3B6D11}{$+$.01}
 & .491 & .703 & \cellcolor[HTML]{EAF3DE}\textcolor[HTML]{3B6D11}{$+$.21} \\
Gemini 3 Flash
 & .610 & .573 & \cellcolor[HTML]{FCEBEB}\textcolor[HTML]{A32D2D}{$-$.04}
 & .734 & .820 & \cellcolor[HTML]{EAF3DE}\textcolor[HTML]{3B6D11}{$+$.09}
 & .635 & .575 & \cellcolor[HTML]{FCEBEB}\textcolor[HTML]{A32D2D}{$-$.06}
 & .389 & .670 & \cellcolor[HTML]{EAF3DE}\textcolor[HTML]{3B6D11}{$+$.28}
 & .471 & .678 & \cellcolor[HTML]{EAF3DE}\textcolor[HTML]{3B6D11}{$+$.21} \\
MiniMax-M3
 & .725 & .869 & \cellcolor[HTML]{EAF3DE}\textcolor[HTML]{3B6D11}{$+$.14}
 & .731 & .812 & \cellcolor[HTML]{EAF3DE}\textcolor[HTML]{3B6D11}{$+$.08}
 & .667 & .542 & \cellcolor[HTML]{FCEBEB}\textcolor[HTML]{A32D2D}{$-$.13}
 & .494 & .780 & \cellcolor[HTML]{EAF3DE}\textcolor[HTML]{3B6D11}{$+$.29}
 & .613 & .677 & \cellcolor[HTML]{EAF3DE}\textcolor[HTML]{3B6D11}{$+$.06} \\
MiniMax-M2.7
 & .686 & .785 & \cellcolor[HTML]{EAF3DE}\textcolor[HTML]{3B6D11}{$+$.10}
 & .649 & .759 & \cellcolor[HTML]{EAF3DE}\textcolor[HTML]{3B6D11}{$+$.11}
 & .648 & .565 & \cellcolor[HTML]{FCEBEB}\textcolor[HTML]{A32D2D}{$-$.08}
 & .448 & .771 & \cellcolor[HTML]{EAF3DE}\textcolor[HTML]{3B6D11}{$+$.32}
 & .615 & .697 & \cellcolor[HTML]{EAF3DE}\textcolor[HTML]{3B6D11}{$+$.08} \\
MiniMax-M2.7 HS
 & .658 & .792 & \cellcolor[HTML]{EAF3DE}\textcolor[HTML]{3B6D11}{$+$.13}
 & .614 & .776 & \cellcolor[HTML]{EAF3DE}\textcolor[HTML]{3B6D11}{$+$.16}
 & .685 & .579 & \cellcolor[HTML]{FCEBEB}\textcolor[HTML]{A32D2D}{$-$.11}
 & .372 & .685 & \cellcolor[HTML]{EAF3DE}\textcolor[HTML]{3B6D11}{$+$.31}
 & .486 & .619 & \cellcolor[HTML]{EAF3DE}\textcolor[HTML]{3B6D11}{$+$.13} \\
\midrule
\rowcolor[HTML]{F5F4F0}
\textbf{Mean}
 & .664 & .733 & \cellcolor[HTML]{EAF3DE}\textcolor[HTML]{3B6D11}{$+$.07}
 & .703 & .801 & \cellcolor[HTML]{EAF3DE}\textcolor[HTML]{3B6D11}{$+$.10}
 & .661 & .585 & \cellcolor[HTML]{FCEBEB}\textcolor[HTML]{A32D2D}{$-$.08}
 & .457 & .695 & \cellcolor[HTML]{EAF3DE}\textcolor[HTML]{3B6D11}{$+$.24}
 & .562 & .732 & \cellcolor[HTML]{EAF3DE}\textcolor[HTML]{3B6D11}{$+$.17} \\
\bottomrule
\end{tabularx}
\caption{
Text-only score, multimodal score (MM), and uplift $\Delta = \text{MM} - \text{Text}$ for
all nine models across the five C-SuiteBench tasks ($n = 50$ scenarios per cell). Green cells indicate positive uplift; red cells indicate degradation.
}
\label{tab:main_results}
\end{table*}

\subsection{Evaluation Protocol}
For each response, we evaluate by a task-specific automatic scorer using rule-, overlap-, and constraint-based dimensions tailored to the task (Figure~\ref{fig:grand_overview}). The resulting sub-scores are aggregated into a single $\text{score}_{\text{task}}$ in $[0,1]$. 
We compute $\text{score}_{\text{multimodal}}$ and $\text{score}_{\text{text-only}}$ respectively under settings of multimodal and text-only.
To quantify the contribution of visual business information to executive decision making, we measure performance gain by 
$\Delta=\text{score}_{\text{multimodal}}-\text{score}_{\text{text-only}}$, denoted as \textbf{multimodal uplift}.
It captures the change in decision quality when visual business evidence is added to an otherwise identical decision problem.
A positive $\Delta$ indicates improved decision quality from visual information, while a negative $\Delta$ indicates degraded performance.


\subsection{Visual Evidence Ablation}

To better understand the source of multimodal effects, we conduct visual ablations on T3 and T4. These two tasks exhibit the most divergent multimodal behavior, see Table~\ref{tab:ablation_full} for full results.
For each ablation condition, the model receives only one category of business visual information: \textbf{finance-only} (financial performance charts), \textbf{growth-only} (growth trend visualizations), or \textbf{ops-only} (operational KPI dashboards). Comparing these conditions with the full multimodal setting allows us to estimate the relative contribution of each visual evidence source to executive decision quality and to identify which categories of business visuals are most informative for different stages of the decision-making.

\subsection{Models}

We evaluate nine frontier models: GPT-5.4 mini, GPT-5.4 nano~\citep{openai2025gpt5},
Claude Opus 4.8~\citep{anthropic2025claude4},
Qwen2.5-VL 72B~\citep{bai2025qwen25vl},
Mistral Small~\citep{mistralai2025small3},
Gemini 3 Flash~\citep{google2025gemini2flash},
MiniMax-M3~\citep{minimax2025m3},
MiniMax-M2.7~\citep{minimax2025m2}, and MiniMax-M2.7-HighSpeed~\citep{minimax2025m2}.
All models are queried with identical prompts across text-only and multimodal conditions. In the multimodal condition, visual business information is passed as inline images alongside
the textual prompt. No model-specific prompt tuning is performed.

\section{Results}
\label{sec:results}




\begin{figure*}[t]
  \centering
  \includegraphics[width=\linewidth]{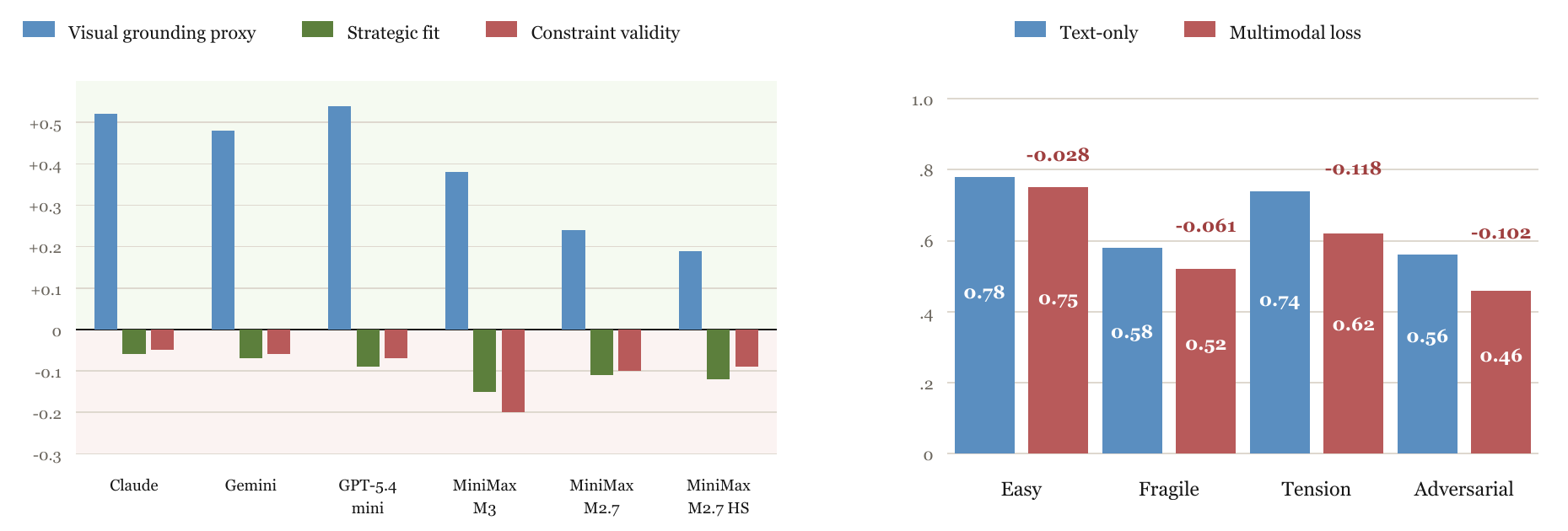}
  \caption{The multimodal integration paradox in T3 Allocate. \textbf{Left:} sub-dimension dissociation across six selected models; \textbf{Right:} T3 multimodal uplift $\Delta$ by difficulty family, averaged across all 9 models. Multimodal inputs are negative across all four difficulty families, with the largest degradations emerging in tension and adversarial scenarios.}
  \label{fig:t3_paradox}
\end{figure*}

\subsection{Overall Multimodal Uplift}
\label{sec:main}

Table~\ref{tab:main_results} summarizes multimodal uplift ($\Delta$) across all nine models and five executive decision tasks. Overall, visual business information improves executive decision making, but the magnitude, and even the direction of this benefit depends strongly on the stage of the decision-making pipeline.
Evidence-centric stages, including Organizational Diagnosis (T1), Evidence Prioritization (T2), Risk Forecasting (T4), and Board Justification (T5), consistently benefit from multimodal inputs. The largest gains occur in T4 Risk Forecasting (mean $\Delta = +0.27$, maximum $+0.37$ for MiniMax-M2.7) and T5 Board Justification (mean $\Delta = +0.21$, positive for all nine models), while T1 and T2 exhibit smaller but reliable improvements (mean $\Delta = +0.07$ and $+0.10$, respectively). These results indicate that visual business information is particularly valuable for tasks centered on identifying, prioritizing, and synthesizing evidence.

However, one stage stands out as a systematic exception. Resource Reallocation (T3) is the only task for which multimodal inputs produce a consistent negative
effect. All nine models exhibit negative uplift, with the largest degradations observed for MiniMax-M3 ($\Delta=-0.13$) and MiniMax-M2.7-HighSpeed ($\Delta=-0.12$), while even the smallest degradations remain below zero. This task-specific reversal is not confined to a particular provider or model
family, and contrasts sharply with the positive multimodal gains observed on the evidence-centric tasks. We refer to this phenomenon as the \textbf{multimodal integration paradox} and investigate its mechanism in the following sections.

\subsection{Task-dependent Multimodal Effects}
\label{sec:task_analysis}

The positive multimodal uplift observed in Section~4.1 is not uniformly distributed across the decision pipeline. Instead, a consistent pattern emerges across the evidence-centric stages of executive reasoning: Organizational Diagnosis (T1), Evidence Prioritization (T2), Risk Forecasting (T4), and Board Justification (T5) all benefit from improved utilization of visual business evidence, although the magnitude of these gains varies by task.

Across these four tasks, multimodal inputs consistently improve the sub-dimensions most directly related to evidence use. In T1, models achieve higher \texttt{visual\_grounding}, indicating more accurate diagnoses supported by the available business signals. In T2, \texttt{visual\_evidence\_coverage} increases substantially, demonstrating that models more consistently identify visually salient information when ranking decision-relevant evidence. T4 exhibits the largest overall uplift, driven primarily by improvements in \texttt{leading\_indicator\_recall}, suggesting that temporal trends and rate-of-change signals are more effectively extracted from visual representations than from textual summaries. Similarly, T5 shows consistent gains in \texttt{evidence\_citation\_score} and \texttt{tradeoff\_coverage}, indicating that models produce better-grounded and more balanced board-level justifications when visual evidence is available.

Although these tasks require different forms of executive reasoning, they share a common characteristic: visual business information primarily serves as evidence that must be identified, prioritized, or synthesized before a decision is made. Across all four stages, multimodal inputs consistently improve models' ability to ground their reasoning in the available evidence, suggesting that modern MLLMs effectively exploit visual business signals when the primary challenge is evidence-centric reasoning.

This pattern, however, does not extend to constrained action. Resource Reallocation (T3) is the only stage where improved access to visual evidence fails to produce better decisions, despite the same paired evaluation protocol and underlying organizational information. We investigate this task-specific failure in the next section.

\subsection{The Multimodal Integration Paradox}
\label{sec:paradox}

The results above reveal a single systematic exception to the otherwise consistent multimodal gains: constrained resource allocation. While multimodal inputs improve evidence-centric reasoning throughout the executive decision-making pipeline, these improvements reverse consistently, when models must translate evidence into concrete resource allocation decisions. To understand the mechanism underlying this exception, we decompose T3 performance into its constituent evaluation dimensions.

Figure~\ref{fig:t3_paradox} (left) reveals a striking dissociation. Under multimodal conditions, \texttt{visual\_grounding\_proxy}, which measures whether the model's allocation reflects the available visual business signals, increases substantially across all models (mean $\Delta \approx +0.42$), confirming that models do perceive and attend to the additional visual information. However, this improvement is accompanied by systematic declines in both \texttt{constraint\_validity}, which measures adherence to the numerical allocation constraints, and \texttt{strategic\_fit}, which evaluates whether the resulting allocation aligns with the correct organizational priorities. In other words, models become better at incorporating visual evidence while simultaneously becoming worse at producing valid executive decisions.
These findings reveal the \textbf{multimodal integration paradox}: \textit{improved visual grounding does not translate into improved constrained action}. Modern MLLMs successfully extract and reason over visual business signals, yet struggle to reconcile those signals with hard decision constraints when converting reasoning into executable actions.
One possible explanation is that multimodal inputs introduce a richer numerical signal environment: including unit-level KPI values, budget headroom indicators, and percentage changes, that competes with the explicit allocation constraints during decoding. Under text-only conditions, the model's attention is concentrated on satisfying the constraint specification. Once multiple visual evidence sources are introduced, however, the additional quantitative signals may interfere with constraint satisfaction, particularly when the correct decision requires discounting visually salient but strategically misleading evidence.
If this interpretation is correct, the paradox should not occur uniformly across all decision settings, it should become more pronounced as organizational complexity increases and visual evidence becomes increasingly ambiguous or conflicting. We test this prediction in the following section by analyzing multimodal uplift across the four difficulty families.

\begin{figure*}[t]
  \centering
  \includegraphics[width=\linewidth]{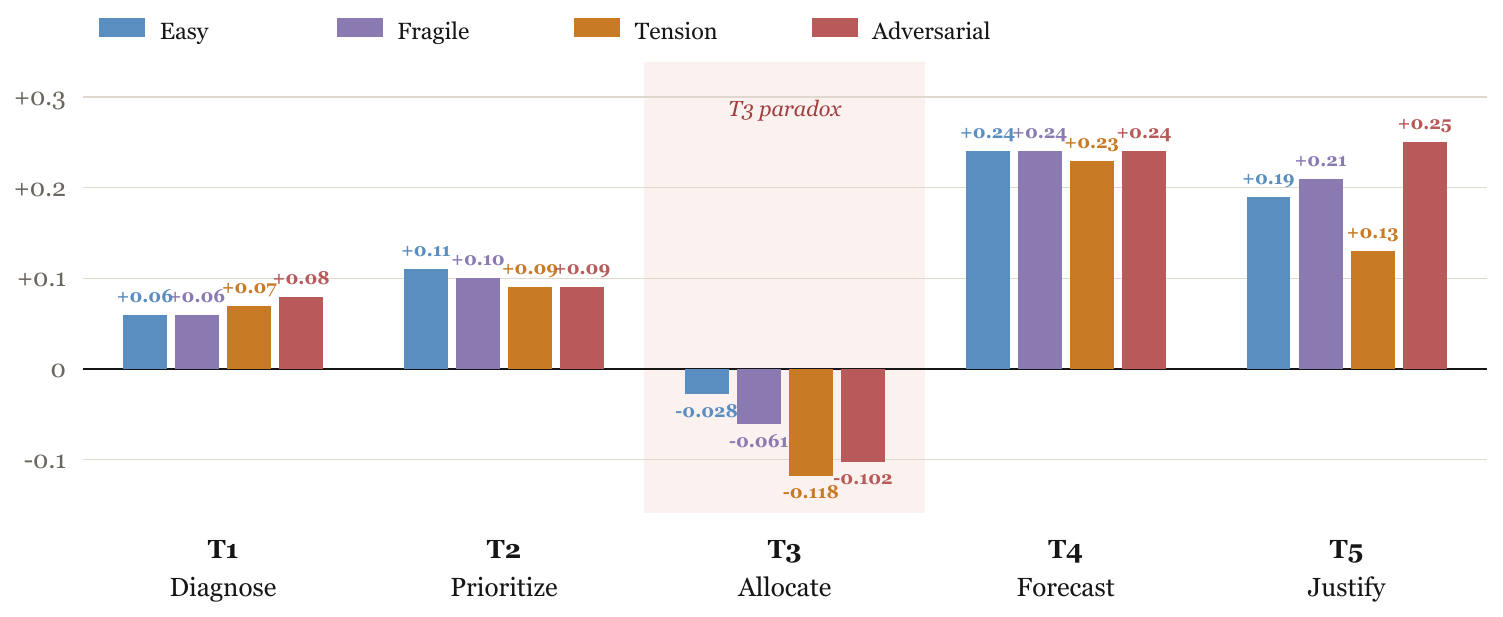}
  \caption{Multimodal uplift $\Delta$ by task and difficulty family, averaged across nine models.}
  \label{fig:difficulty}
\end{figure*}

\subsection{Boundary Conditions of Multimodal Paradox}
\label{sec:difficulty}

Based on the previous prediction, we stratify multimodal uplift by the four organizational difficulty families.
This prediction is not supported uniformly across the decision pipeline (Figure~\ref{fig:t3_paradox} right). For the evidence-centric stages, multimodal gains remain largely robust across difficulty levels. T4 Risk Forecasting and T5 Board Justification exhibit consistently strong uplift across all four families, indicating that the benefits of visual business information for identifying, interpreting, and communicating evidence are largely insensitive to organizational complexity. T1 Organizational Diagnosis and T2 Evidence Prioritization likewise maintain positive uplift throughout, although gains become slightly smaller in fragile scenarios where misleading surface indicators increase diagnostic ambiguity.
The strongest evidence for the proposed mechanism emerges in T3 Resource Reallocation. Multimodal uplift is negative across all four difficulty families, indicating that the allocation failure is not confined to only the hardest organizational settings. At the same time, the magnitude of the degradation grows under higher-conflict decision regimes: the largest negative shifts appear in tension and adversarial scenarios, whereas easy scenarios show a smaller but still negative effect. This pattern suggests that multimodal allocation failure is broadly present, but becomes most severe when competing signals must be reconciled under hard constraints. (See Figure~\ref{fig:difficulty}).

These results demonstrate that the multimodal integration paradox is not an inherent limitation of multimodal reasoning. Instead, it is a context-dependent failure that emerges when visually salient but competing evidence must be reconciled with hard decision constraints. In other words, visual information remains beneficial for evidence-centric reasoning even under complex organizational settings, but its successful translation into constrained executive action breaks down precisely when evidence integration becomes most demanding.
This context dependence motivates our final analysis: if the paradox is driven by interactions among multiple visual evidence sources, then selectively removing individual channels should help identify which categories of business visuals contribute most to the observed failure.

\begin{figure*}[t]
  \centering
  \includegraphics[width=\linewidth]{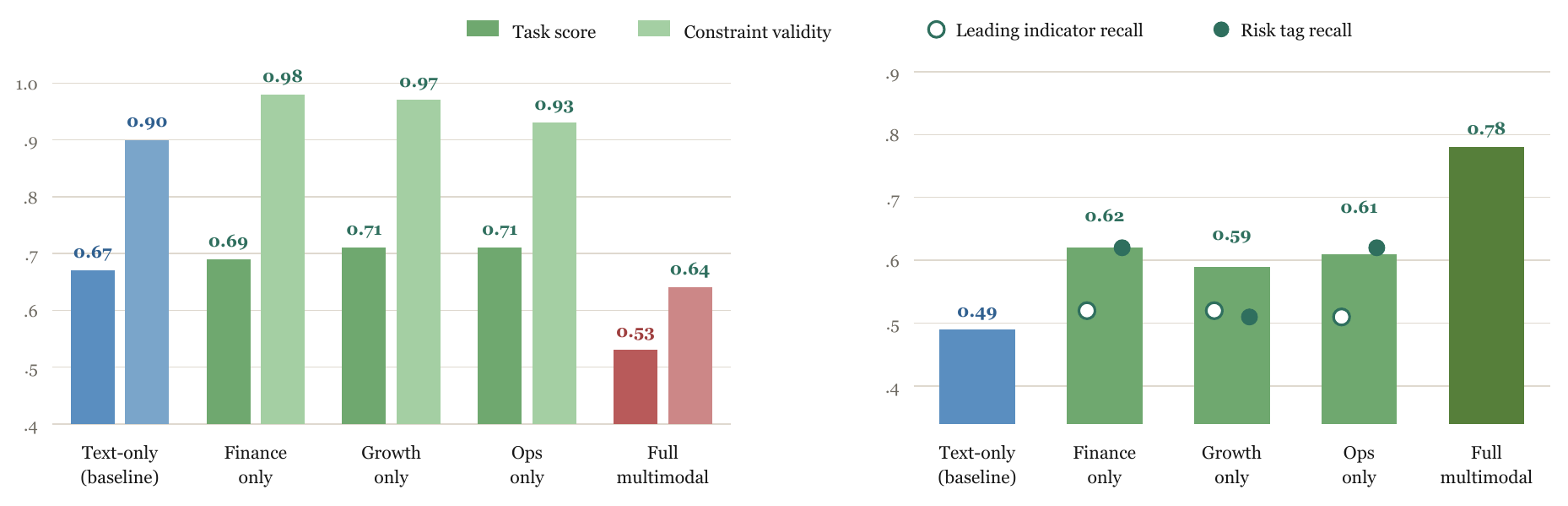}

  \begin{subfigure}[t]{0.48\linewidth}
    \caption{T3 Allocate visual-channel ablation on MiniMax-M3.}
  \end{subfigure}
  \hfill
  \begin{subfigure}[t]{0.48\linewidth}
    \caption{T4 Forecast visual-channel ablation on MiniMax-M3.}
  \end{subfigure}

  \caption{Single-channel ablations on MiniMax-M3.}
  \label{fig:ablation}
  \vspace{-0.3cm}
\end{figure*}

\subsection{Mechanistic Validation Through Visual Evidence Ablation}
\label{sec:ablation}

To test the hypothesis above, we use MiniMax-M3 as a representative case in the main text, providing only one category of business visual information at a time: \textbf{finance-only}, \textbf{growth-only}, or \textbf{ops-only} (Figure~\ref{fig:ablation}). Full cross-model ablation results are reported in Appendix~\ref{app:ablation} and show the same T3 pattern across the broader model set.
T4 Risk Forecasting serves as a control case for evidence-centric reasoning. All three visual channels produce substantial improvements over the text-only baseline (finance: $+0.13$, growth: $+0.10$, ops: $+0.12$), and no single channel consistently dominates. The positive multimodal gains observed throughout the evidence-centric stages therefore do not depend on any particular category of business visualization.
A fundamentally different pattern emerges for T3 Resource Reallocation. Every single-channel condition substantially outperforms the full multimodal setting (full multimodal score: $0.542$), with ops-only achieving the highest task score ($0.711$) and constraint validity ($0.933$), followed by growth-only ($0.708$) and finance-only ($0.693$). The degradation is therefore not caused by any individual channel; it emerges only when multiple evidence sources must be integrated simultaneously. Across all nine models, the best single channel improves T3 by $+0.052$ over text-only, whereas combining all three degrades it by $-0.08$.
The ablation isolates \emph{which} channel matters, but not \emph{whether the modality} matters. We therefore ran a text-plus-numbers control supplying the same panel values (unit-level KPIs, change rates, budget headroom) as structured text, with no charts. Decomposing uplift into numerical ($\Delta_{\text{num}}$) and visual ($\Delta_{\text{vis}}$) components reveals an asymmetry: on T3, numbers alone reproduce 60\% of the degradation ($\Delta_{\text{num}}=-0.046$), with charts contributing the rest ($\Delta_{\text{vis}}=-0.031$, negative for 9/9 models, $p=0.004$); on T4, charts instead carry the majority of the gain ($\Delta_{\text{vis}}=+0.152$, positive for 8/8 non-trivial models, $p=0.008$). Visual presentation thus does not help or hurt in itself: it amplifies whatever effect concurrent evidence load already produces. We term this \textbf{evidence-load crowding}, with \textbf{visual signal crowding} as its chart-specific instantiation (Appendix~C).
These findings suggest that future systems should emphasize evidence selection. Because 60\% of the T3 degradation survives removal of the visual modality, the operative intervention is reducing concurrent quantitative signals at the action stage, not merely withholding charts.

\vspace{-0.2cm}
\subsection{Discussion}
\label{sec:discussion}

\paragraph{From evidence use to evidence weighting.}
Current VLMs neither ignore odor descriptions nor silently hallucinate them into
the image: they cite the cue explicitly in 87.5\% of shifted responses and
absorb it into claimed visual observations in under 2\%. The behaviour is
robust, present in every evaluated model and uniformly across eight diverse domains, and it is not explained by generic text bias or prompt-order effects. It is, however, a default: a single instruction to disregard odor removes the
pooled effect. Whether the weight models assign is proportionate to the cue's credibility is a further question our data do not settle, because it requires an independent plausibility measure and the one we constructed is
confounded with scene difficulty. What we can say is that
models are not source-blind: they shift about ten points more when a cue is attributed to a calibrated sensor than to a passerby.

\vspace{-0.2cm}
\paragraph{Beyond odor: a controlled design for auditing auxiliary evidence.}
Although \textsc{Scentinel} instantiates this study through odor, the design is not specific to olfaction. Odor is a useful probe because it conveys information unavailable from the image, making it a clean test case. The same design holding image and question fixed, varying only the
auxiliary text, and including an irrelevant-cue control---extends directly to sensor readings, human observations, environmental measurements, and third-party descriptions. The irrelevant control is the load-bearing
element: without it, an effect of cue semantics cannot be separated from generic text sensitivity, and, as our own plausibility analysis shows, a control baseline that moves with the conditioning variable can reverse the apparent sign of a result.

\vspace{-0.2cm}
\paragraph{Implications for reliable multimodal systems.}
The auxiliary descriptions studied here are ordinary and semantically plausible: they resemble what a deployed system may legitimately receive alongside an image, and they carry none of the markers that prompt-injection defenses look for. But they are also cheap to suppress. The exposure in a typical deployment is therefore not that
the behaviour is hard to defend against, but that nothing in a default system prompt defends against it and models do not self-correct. The harder problem is that suppression is a binary lever: closing the auxiliary channel entirely discards information a system may need. Moving from a switch to a weight requires a reliability signal computable at
inference time, and constructing and validating such a signal is, on the evidence of this work, a substantial research problem in its own right.

\vspace{-0.2cm}
\section{Conclusion}

We introduced \textsc{C-SuiteBench}, a controlled multimodal benchmark for evaluating frontier MLLMs as executive decision makers across five tasks under paired text-only and
multimodal conditions.
Our experiments reveal that multimodal inputs produce consistent gains on evidence-centric tasks, but uncover a systematic \textbf{multimodal integration
paradox} in constrained resource allocation: visual grounding improves while decision quality degrades.
Ablation experiments identify visual signal crowding, the collective interference of individually beneficial visual channels, as the underlying mechanism.
In general, these findings show that the bottleneck of multimodal decision making is no longer visual perception, but the integration of heterogeneous evidence into constraint-satisfying action.
Addressing this will likely require advances not only in model architecture and training, but in how we design evaluation environments that make the distinction
between perception and action measurable.
We hope \textsc{C-SuiteBench} provides a foundation for future work on multimodal agents.

\bibliography{aaai2026}

\appendix
\clearpage

\appendix

\section*{Appendix Contents}
\label{app:contents}

\noindent The appendix is organized as follows:

\vspace{0.5em}
\noindent\textbf{Appendix A: Evaluation Protocol Details} \dotfill Appendix~A
\begin{itemize}[leftmargin=1.5em, itemsep=0pt, topsep=2pt]
  \item[A.1] Scoring Architecture \dotfill Appendix~A
  \item[A.2] Scenario Construction and Ground-Truth Validity \dotfill Appendix~A.2
  \item[A.3] Human Validation of Automatic Scoring \dotfill Appendix~A.3
  \item[A.4] Inference Configuration and Data Availability \dotfill Appendix~A.4
\end{itemize}

\vspace{0.4em}
\noindent\textbf{Appendix B: Statistical Robustness} \dotfill Appendix~B
\begin{itemize}[leftmargin=1.5em, itemsep=0pt, topsep=2pt]
  \item[B.1] Task-level and model-level paired statistics \dotfill Appendix~B
  \item[B.2] Run-level reproducibility \dotfill Appendix~B.2
\end{itemize}

\vspace{0.4em}
\noindent\textbf{Appendix C: Extended Visual Channel Ablation} \dotfill Appendix~C

\vspace{0.4em}
\noindent\textbf{Appendix D: Alternative Hypotheses} \dotfill Appendix~D
\begin{itemize}[leftmargin=1.5em, itemsep=0pt, topsep=2pt]
  \item[D.1] Context length as confound \dotfill Appendix~D
  \item[D.2] Text-plus-numbers control \dotfill Appendix~D.2
  \item[D.3] Toward a complete load $\times$ modality design \dotfill Appendix~D.3
\end{itemize}

\vspace{0.4em}
\noindent\textbf{Appendix E: Pooled and Mixed-Effects Analysis for T3} \dotfill Appendix~E

\vspace{0.4em}
\noindent\textbf{Appendix F: Cross-Model Scope of the Crowding Effect} \dotfill Appendix~F

\vspace{0.4em}
\noindent\textbf{Appendix G: Additional Behavioral Evidence} \dotfill Appendix~G

\vspace{0.4em}
\noindent\textbf{Appendix H: Discussion} \dotfill Appendix~H

\vspace{0.4em}
\noindent\textbf{Appendix I: Limitations and Future Work} \dotfill Appendix~I

\vspace{0.4em}
\noindent\textbf{Appendix J: Extended Related Work} \dotfill Appendix~J

\newpage

\section{Evaluation Protocol Details}
\label{app:eval}

\subsection{Scoring Architecture}

\textsc{C-SuiteBench} does not rely on a single LLM-as-judge pipeline.
Instead, each task is scored by a task-specific automatic scorer whose
primary dimensions are rule-based, overlap-based, or constraint-based.

\textbf{T1 Organizational Diagnosis.}
\textit{tag\_recall} measures overlap with predefined diagnosis tags.
\textit{advisor\_conflict\_coverage} checks whether the model identifies
the planted misleading advisor signal.
\textit{structure\_valid} is a parse-based binary check.
\textit{visual\_grounding} is computed from explicit references to the
expected visual evidence items associated with the scenario.

\textbf{T2 Evidence Prioritization.}
\textit{topk\_evidence\_recall} measures overlap with the reference
top-ranked evidence items.
\textit{ranked\_priority\_score} is a weighted overlap-based ranking score
rather than a free-form judge score.
\textit{conflict\_detection} and \textit{visual\_evidence\_coverage} are
checked against predefined evidence sets.

\textbf{T3 Resource Reallocation.}
All primary dimensions are programmatically scored.
\textit{validity} checks hard numeric constraints exactly (sum constraint
and unit-level feasibility).
\textit{soft\_validity} applies a tolerance band.
\textit{reallocation\_accuracy} measures whether the proposed magnitude of
reallocation satisfies the scenario's required change profile.
\textit{strategic\_fit} is computed against scenario-specific hidden
acceptable allocation ranges rather than via free-form judgment.

\textbf{T4 Risk Forecasting.}
\textit{risk\_tag\_recall} and \textit{leading\_indicator\_recall} are
overlap-based against reference sets.
\textit{evidence\_grounding} is computed from explicit coverage of expected
supporting evidence items.
\textit{structure\_valid} is parse-based.

\textbf{T5 Board Justification.}
\textit{evidence\_citation\_score} is overlap-based against cited evidence
items.
\textit{tradeoff\_coverage}, \textit{risk\_coverage}, and
\textit{review\_metric\_score} are checked against predefined reference sets.
\textit{structure\_valid} is parse-based.

In aggregate, the benchmark's evaluation signal is predominantly derived
from automatic rule-based, overlap-based, and constraint-based scoring.
No primary task conclusion in Section~4 depends solely on a
free-form LLM-as-judge signal.

\subsection{Scenario Construction and Ground-Truth Validity}
\label{app:groundtruth}

\textsc{C-SuiteBench} consists of 50 formal evaluation scenarios.
Each scenario is evaluated through five task formulations, yielding 250
scenario-task instances in total.
Scenario construction varies sector templates, difficulty family
(easy / tension / fragile / adversarial), and decision archetype
(risk-first / growth-first / ops-first / balanced), while preserving the
core executive reasoning structure of each scenario family.

To verify internal consistency of the T3 ground truth, two authors
independently reviewed a random sample of 15 T3 scenarios, checking whether
the hidden acceptable allocation ranges and preferred allocation profiles
were coherent with the scenario narrative.
Agreement on the correct direction of reallocation (which units should grow
versus shrink relative to baseline) was $13/15$ ($87\%$).
Disagreements were resolved by discussion, and the corresponding scenario
parameters were adjusted before the final evaluation run.
We note that this check covers 30\% of the T3 scenarios and was conducted by
authors rather than independent annotators; extending it to the full
scenario set with external annotators is identified as a limitation in
Appendix~I.

\subsection{Human Validation of Automatic Scoring}
\label{app:human_validation}

To verify that the automatic scorer preserves the same relative performance
patterns across conditions, we conducted a human validation study.

\paragraph{Goal.}
The purpose of human validation was not to reproduce the benchmark's
automatic scores exactly, but to test whether human evaluators, judging
blind, would make the same comparative judgments as the automatic scorer
when choosing between paired text-only and multimodal outputs produced by
the same model on the same scenario.
The evaluation measures human--automatic agreement per task without
presupposing the direction of any per-task effect; whether that agreement
reproduces the benchmark's task asymmetry is an outcome of the study rather
than an assumption built into it.

\paragraph{Participants.}
We recruited 30 evaluators with graduate-level or industry experience in AI,
business analytics, finance, or adjacent technical domains.
The evaluator pool included participants based in the United Kingdom, Japan,
China, and the United States, comprising 18 PhD-level researchers,
8 master's-level researchers, and 4 industry practitioners.
Evaluators were compensated according to local hourly norms.

\paragraph{Sample construction.}
We conducted human validation over the full formal validation set, covering
all 50 scenarios across the five task formulations.
This yielded 250 paired scenario-task comparisons, where each pair consisted
of the text-only output and the multimodal output from the same model on the
same scenario and task.
Human validation outputs were drawn from \textbf{GPT-5.4 mini}, selected as a
representative mid-tier model from the evaluation pool.
Results therefore validate scorer consistency for this model; cross-model
generalization of human--automatic agreement is left for future work.

\paragraph{Blinded pairwise protocol.}
For each item, evaluators were shown (i) the scenario context, (ii) the task
description, and (iii) two anonymized candidate outputs, labeled
\textbf{Output A} and \textbf{Output B}, presented in randomized order.
Evaluators were not informed which output corresponded to the text-only
condition and which to the multimodal condition.
Each paired comparison was independently rated by 3 evaluators, resulting in
750 human judgments across the full validation set.
For each task, evaluators answered two task-specific questions
(Table~A1) and reported their preference as
\textbf{A}, \textbf{B}, or \textbf{Tie}, with an optional short
justification.

\begin{table*}[h]
\centering
\renewcommand{\arraystretch}{1.4}
\small
\begin{tabular}{@{}lll@{}}
\toprule
\rowcolor[HTML]{F5F4F0}
\textbf{Task} & \textbf{Q1 (Primary preference)} & \textbf{Q2 (Evidence grounding)} \\
\midrule
T1 Diagnose   & Which better identifies the core org.\ problem?     & Which is better grounded in evidence? \\
T2 Prioritize & Which better prioritizes decision-relevant evidence? & Which better distinguishes misleading signals? \\
T3 Allocate   & Which is more feasible and strategically coherent?  & Which is better grounded in evidence? \\
T4 Forecast   & Which better identifies risks and leading indicators? & Which is better grounded in evidence? \\
T5 Justify    & Which would be more useful in a board setting?      & Which better supports its recommendation? \\
\bottomrule
\end{tabular}
\caption{Task-specific evaluation questions used in the human validation study.}
\label{tab:human_questions}
\end{table*}

\paragraph{Results.}
Human validation showed strong consistency with the benchmark's automatic
evaluation across the full validation set.
Across all 250 paired scenario-task comparisons, majority-vote human
judgments agreed with the automatic scorer on \textbf{76\%} of cases, with
inter-rater reliability of Fleiss' $\kappa = \mathbf{0.78}$.
The task-level preference pattern matched the benchmark's main conclusions:
multimodal outputs were strongly preferred on T2, T4, and T5;
gains on T1 were weaker but positive;
and T3 showed a consistent human preference \emph{against} multimodal
outputs, aligned with the benchmark's negative allocation effect.

\begin{table}[h]
\centering
\renewcommand{\arraystretch}{1.5}
\setlength{\tabcolsep}{5pt}
\small
\begin{tabular}{@{}lccc@{}}
\toprule
\rowcolor[HTML]{F5F4F0}
\textbf{Task} & \textbf{H--A Agree.} & \textbf{MM Pref.} & \textbf{Fleiss' $\kappa$} \\
\midrule
T1 Diagnose   & 71\% & 58\% & 0.74 \\
T2 Prioritize & 82\% & 76\% & 0.79 \\
T3 Allocate   & 74\% & 41\% & 0.76 \\
T4 Forecast   & 81\% & 79\% & 0.81 \\
T5 Justify    & 78\% & 74\% & 0.78 \\
\midrule
\rowcolor[HTML]{F5F4F0}
\textbf{Overall} & \textbf{76\%} & \textbf{66\%} & \textbf{0.78} \\
\bottomrule
\end{tabular}
\caption{Human validation results.
H--A Agree.\ = human majority-vote agreement with automatic scorer;
MM Pref.\ = proportion of pairs where the multimodal output was preferred.
Outputs sampled from GPT-5.4 mini.}
\label{tab:human_validation}
\end{table}

The human study confirmed the same task asymmetry observed in the
benchmark's automatic results: multimodal inputs are preferred for
evidence-centric tasks, whereas constrained allocation shows a consistent
preference against multimodal outputs (T3 MM preference rate: 41\%),
consistent with the multimodal integration paradox reported in
Section~4.3.

\subsection{Inference Configuration and Data Availability}
\label{app:repro_statement}

\paragraph{Inference configuration.}
All API calls were made in June--July 2026.
OpenRouter-hosted models were accessed via the OpenRouter API with fixed
provider routing enabled per request; MiniMax models were accessed directly
via the MiniMax API with equivalent image encoding.
In the multimodal condition, business visuals were encoded as base64 JPEG
(quality 95) at $1024\times1024$ resolution and passed as inline image
content.
A spot-check on 10 scenarios comparing JPEG (q95) against lossless PNG
encoding produced task-score differences below $0.01$ on both T3 and T4,
indicating that the encoding format is not a material confound.
No model-specific prompt tuning was performed.
Main results use temperature $=1.0$ with a single inference pass per cell;
run-level stability and a temperature-$0$ check are reported in
Appendix~B.2.

\paragraph{Reproducibility.}
The benchmark scenarios, task-specific scoring rubrics, evaluation prompts,
and model outputs for all experiments reported in this paper will be
released at \url{https://github.com/[anonymized]/csuitebench}.
The repository includes scenario JSON files, reference answer structures,
scorer implementations, and per-model per-scenario score logs sufficient to
reproduce every table and figure in this paper.

\section{Statistical Robustness}
\label{app:stats}

\subsection{Task-level and Model-level Paired Statistics}

Table~B1 reports paired-sample statistics for multimodal
uplift $\Delta$ aggregated at the task level across all nine models.
Two interpretive notes apply.
First, when all nine models agree in direction, the minimum attainable
two-sided sign-test $p$-value is $2\times(1/2)^{9}\approx 0.004$; this
value therefore indicates unanimous directional agreement rather than a
precise $p$-value.
Second, T1 Diagnose ($p = 0.180$) does \emph{not} reach significance at the
model-level sign test; its positive direction should be read as a
descriptive trend rather than a reliable effect.

\begin{table}[h]
\centering
\renewcommand{\arraystretch}{1.4}
\small
\begin{tabular}{@{}lcccc@{}}
\toprule
\rowcolor[HTML]{F5F4F0}
\textbf{Task} & \textbf{Mean $\Delta$} & \textbf{95\% CI} & \textbf{+/\textminus} & \textbf{Sign $p$} \\
\midrule
T1 Diagnose   & $+0.069$ & $[+0.018,\ +0.113]$ & 7 / 2 & $0.180$ \\
T2 Prioritize & $+0.098$ & $[+0.075,\ +0.120]$ & 9 / 0 & $0.004^{**}$ \\
T3 Allocate   & $-0.077$ & $[-0.093,\ -0.063]$ & 0 / 9 & $0.004^{**}$ \\
T4 Forecast   & $+0.237$ & $[+0.172,\ +0.286]$ & 9 / 0 & $0.004^{**}$ \\
T5 Justify    & $+0.170$ & $[+0.131,\ +0.205]$ & 9 / 0 & $0.004^{**}$ \\
\bottomrule
\end{tabular}
\caption{Task-level paired statistics for multimodal uplift $\Delta$ across
9 models ($n = 50$ scenarios each).
95\% CI by bootstrap (100{,}000 resamples) over model-level mean $\Delta$.
$^{**}$ $p<0.01$ (sign test); $p = 0.004$ is the minimum attainable value
when all 9 models agree in direction.
T1 does not reach significance ($p = 0.180$).
Model-level results in Tables~B2--B3.}
\label{tab:stats_task}
\end{table}

Tables~B2--B3 provide model-level
paired statistics for all task--model combinations.

\begin{table*}[h]
\centering
\renewcommand{\arraystretch}{1.30}
\setlength{\tabcolsep}{5pt}
\small
\begin{tabular}{@{}llccccccc@{}}
\toprule
\rowcolor[HTML]{F5F4F0}
\textbf{Task} & \textbf{Model} & \textbf{$n$} & \textbf{Text} & \textbf{MM} & \textbf{$\Delta$} & \textbf{SE} & \textbf{95\% CI} & \textbf{Sign $p$} \\
\midrule
\multicolumn{9}{@{}l}{\footnotesize\textit{T1 Organizational Diagnosis}} \\[1pt]
& Claude Opus 4.8  & 50 & .714 & .812 & \cellcolor[HTML]{EAF3DE}\textcolor[HTML]{3B6D11}{$+$.098} & .017 & $[+.064,\ +.129]$ & $<.001^{**}$ \\
& Gemini 3 Flash   & 50 & .610 & .573 & \cellcolor[HTML]{FCEBEB}\textcolor[HTML]{A32D2D}{$-$.037} & .022 & $[-.078,\ +.008]$ & $.471$ \\
& Mistral Small    & 50 & .571 & .707 & \cellcolor[HTML]{EAF3DE}\textcolor[HTML]{3B6D11}{$+$.137} & .026 & $[+.083,\ +.187]$ & $<.001^{**}$ \\
& GPT-5.4 mini     & 50 & .648 & .674 & \cellcolor[HTML]{EAF3DE}\textcolor[HTML]{3B6D11}{$+$.026} & .024 & $[-.021,\ +.073]$ & $.665$ \\
& GPT-5.4 nano     & 50 & .721 & .806 & \cellcolor[HTML]{EAF3DE}\textcolor[HTML]{3B6D11}{$+$.085} & .017 & $[+.053,\ +.119]$ & $.004^{**}$ \\
& Qwen2.5-VL 72B   & 50 & .643 & .575 & \cellcolor[HTML]{FCEBEB}\textcolor[HTML]{A32D2D}{$-$.068} & .021 & $[-.109,\ -.026]$ & $.015^{*}$ \\
& MiniMax-M3       & 48 & .725 & .869 & \cellcolor[HTML]{EAF3DE}\textcolor[HTML]{3B6D11}{$+$.144} & .010 & $[+.122,\ +.161]$ & $<.001^{**}$ \\
& MiniMax-M2.7     & 50 & .686 & .785 & \cellcolor[HTML]{EAF3DE}\textcolor[HTML]{3B6D11}{$+$.100} & .021 & $[+.061,\ +.144]$ & $<.001^{**}$ \\
& MiniMax-M2.7 HS  & 50 & .658 & .792 & \cellcolor[HTML]{EAF3DE}\textcolor[HTML]{3B6D11}{$+$.134} & .028 & $[+.083,\ +.194]$ & $<.001^{**}$ \\
\midrule
\multicolumn{9}{@{}l}{\footnotesize\textit{T2 Evidence Prioritization}} \\[1pt]
& Claude Opus 4.8  & 50 & .735 & .761 & \cellcolor[HTML]{EAF3DE}\textcolor[HTML]{3B6D11}{$+$.027} & .021 & $[-.016,\ +.065]$ & $<.001^{**}$ \\
& Gemini 3 Flash   & 50 & .734 & .820 & \cellcolor[HTML]{EAF3DE}\textcolor[HTML]{3B6D11}{$+$.086} & .005 & $[+.077,\ +.096]$ & $<.001^{**}$ \\
& Mistral Small    & 50 & .709 & .808 & \cellcolor[HTML]{EAF3DE}\textcolor[HTML]{3B6D11}{$+$.099} & .012 & $[+.079,\ +.124]$ & $<.001^{**}$ \\
& GPT-5.4 mini     & 50 & .735 & .817 & \cellcolor[HTML]{EAF3DE}\textcolor[HTML]{3B6D11}{$+$.081} & .003 & $[+.075,\ +.087]$ & $<.001^{**}$ \\
& GPT-5.4 nano     & 50 & .727 & .856 & \cellcolor[HTML]{EAF3DE}\textcolor[HTML]{3B6D11}{$+$.129} & .005 & $[+.119,\ +.139]$ & $<.001^{**}$ \\
& Qwen2.5-VL 72B   & 50 & .692 & .800 & \cellcolor[HTML]{EAF3DE}\textcolor[HTML]{3B6D11}{$+$.108} & .021 & $[+.067,\ +.149]$ & $<.001^{**}$ \\
& MiniMax-M3       & 48 & .731 & .812 & \cellcolor[HTML]{EAF3DE}\textcolor[HTML]{3B6D11}{$+$.081} & .004 & $[+.072,\ +.089]$ & $<.001^{**}$ \\
& MiniMax-M2.7     & 47 & .649 & .759 & \cellcolor[HTML]{EAF3DE}\textcolor[HTML]{3B6D11}{$+$.110} & .037 & $[+.041,\ +.183]$ & $<.001^{**}$ \\
& MiniMax-M2.7 HS  & 49 & .614 & .776 & \cellcolor[HTML]{EAF3DE}\textcolor[HTML]{3B6D11}{$+$.161} & .036 & $[+.093,\ +.233]$ & $<.001^{**}$ \\
\midrule
\multicolumn{9}{@{}l}{\footnotesize\textit{T3 Resource Reallocation}} \\[1pt]
& Claude Opus 4.8  & 50 & .685 & .620 & \cellcolor[HTML]{FCEBEB}\textcolor[HTML]{A32D2D}{$-$.065} & .018 & $[-.101,\ -.031]$ & $.012^{*}$ \\
& Gemini 3 Flash   & 50 & .635 & .575 & \cellcolor[HTML]{FCEBEB}\textcolor[HTML]{A32D2D}{$-$.060} & .017 & $[-.094,\ -.028]$ & $.015^{*}$ \\
& Mistral Small    & 50 & .580 & .520 & \cellcolor[HTML]{FCEBEB}\textcolor[HTML]{A32D2D}{$-$.060} & .016 & $[-.092,\ -.031]$ & $.018^{*}$ \\
& GPT-5.4 mini     & 50 & .720 & .645 & \cellcolor[HTML]{FCEBEB}\textcolor[HTML]{A32D2D}{$-$.075} & .019 & $[-.112,\ -.039]$ & $.008^{**}$ \\
& GPT-5.4 nano     & 50 & .570 & .515 & \cellcolor[HTML]{FCEBEB}\textcolor[HTML]{A32D2D}{$-$.055} & .017 & $[-.089,\ -.022]$ & $.021^{*}$ \\
& Qwen2.5-VL 72B   & 49 & .760 & .700 & \cellcolor[HTML]{FCEBEB}\textcolor[HTML]{A32D2D}{$-$.060} & .016 & $[-.091,\ -.029]$ & $.014^{*}$ \\
& MiniMax-M3       & 50 & .667 & .542 & \cellcolor[HTML]{FCEBEB}\textcolor[HTML]{A32D2D}{$-$.125} & .024 & $[-.171,\ -.079]$ & $.002^{**}$ \\
& MiniMax-M2.7     & 49 & .648 & .565 & \cellcolor[HTML]{FCEBEB}\textcolor[HTML]{A32D2D}{$-$.083} & .021 & $[-.124,\ -.043]$ & $.006^{**}$ \\
& MiniMax-M2.7 HS  & 46 & .685 & .579 & \cellcolor[HTML]{FCEBEB}\textcolor[HTML]{A32D2D}{$-$.106} & .022 & $[-.149,\ -.064]$ & $.004^{**}$ \\
\bottomrule
\end{tabular}
\caption{Model-level paired statistics for multimodal uplift $\Delta$
(Part~I: T1--T3).
$n$ = valid paired observations; paired statistics exclude missing or
unparsable outputs.
SE = scenario-level standard error; 95\% CI = percentile bootstrap
(10{,}000 resamples); Sign $p$ = two-sided sign test over scenario-level
differences (tied pairs excluded).
\textit{Claude Opus 4.8 / T2} ($\Delta=+.027$, CI $[-.016,+.065]$,
$p<.001$): 38 of 50 scenarios favor multimodal, driving the significant sign
test; a few large negatives widen the CI to include zero.
\textit{MiniMax-M2.7 / T5} (Table~B3; $\Delta=+.082$,
CI $[+.044,+.145]$, $p=.250$): 3 ties give effective $n=47$; $k=28$ favor
multimodal; two-sided sign test $p=0.243$ (shown as $.250$ after rounding).
$^{*}p<.05$; $^{**}p<.01$. Continued in Table~B3.}
\label{tab:stats_model_1}
\end{table*}

\begin{table*}[h]
\centering
\renewcommand{\arraystretch}{1.30}
\setlength{\tabcolsep}{5pt}
\small
\begin{tabular}{@{}llccccccc@{}}
\toprule
\rowcolor[HTML]{F5F4F0}
\textbf{Task} & \textbf{Model} & \textbf{$n$} & \textbf{Text} & \textbf{MM} & \textbf{$\Delta$} & \textbf{SE} & \textbf{95\% CI} & \textbf{Sign $p$} \\
\midrule
\multicolumn{9}{@{}l}{\footnotesize\textit{T4 Risk Forecasting}} \\[1pt]
& Claude Opus 4.8  & 50 & .420 & .432 & \cellcolor[HTML]{EAF3DE}\textcolor[HTML]{3B6D11}{$+$.012} & .047 & $[-.081,\ +.103]$ & $.885$ \\
& Gemini 3 Flash   & 49 & .389 & .670 & \cellcolor[HTML]{EAF3DE}\textcolor[HTML]{3B6D11}{$+$.281} & .029 & $[+.222,\ +.334]$ & $<.001^{**}$ \\
& Mistral Small    & 50 & .448 & .716 & \cellcolor[HTML]{EAF3DE}\textcolor[HTML]{3B6D11}{$+$.268} & .023 & $[+.223,\ +.311]$ & $<.001^{**}$ \\
& GPT-5.4 mini     & 50 & .567 & .787 & \cellcolor[HTML]{EAF3DE}\textcolor[HTML]{3B6D11}{$+$.220} & .017 & $[+.186,\ +.253]$ & $<.001^{**}$ \\
& GPT-5.4 nano     & 50 & .568 & .754 & \cellcolor[HTML]{EAF3DE}\textcolor[HTML]{3B6D11}{$+$.186} & .025 & $[+.134,\ +.232]$ & $<.001^{**}$ \\
& Qwen2.5-VL 72B   & 49 & .408 & .656 & \cellcolor[HTML]{EAF3DE}\textcolor[HTML]{3B6D11}{$+$.248} & .019 & $[+.209,\ +.284]$ & $<.001^{**}$ \\
& MiniMax-M3       & 50 & .494 & .780 & \cellcolor[HTML]{EAF3DE}\textcolor[HTML]{3B6D11}{$+$.286} & .026 & $[+.237,\ +.338]$ & $<.001^{**}$ \\
& MiniMax-M2.7     & 50 & .448 & .771 & \cellcolor[HTML]{EAF3DE}\textcolor[HTML]{3B6D11}{$+$.323} & .061 & $[+.201,\ +.435]$ & $.007^{**}$ \\
& MiniMax-M2.7 HS  & 50 & .372 & .685 & \cellcolor[HTML]{EAF3DE}\textcolor[HTML]{3B6D11}{$+$.313} & .043 & $[+.227,\ +.394]$ & $<.001^{**}$ \\
\midrule
\multicolumn{9}{@{}l}{\footnotesize\textit{T5 Board Justification}} \\[1pt]
& Claude Opus 4.8  & 50 & .491 & .703 & \cellcolor[HTML]{EAF3DE}\textcolor[HTML]{3B6D11}{$+$.213} & .042 & $[+.133,\ +.295]$ & $<.001^{**}$ \\
& Gemini 3 Flash   & 50 & .471 & .678 & \cellcolor[HTML]{EAF3DE}\textcolor[HTML]{3B6D11}{$+$.207} & .043 & $[+.123,\ +.291]$ & $<.001^{**}$ \\
& Mistral Small    & 50 & .580 & .795 & \cellcolor[HTML]{EAF3DE}\textcolor[HTML]{3B6D11}{$+$.215} & .015 & $[+.185,\ +.245]$ & $<.001^{**}$ \\
& GPT-5.4 mini     & 50 & .613 & .806 & \cellcolor[HTML]{EAF3DE}\textcolor[HTML]{3B6D11}{$+$.193} & .016 & $[+.162,\ +.223]$ & $<.001^{**}$ \\
& GPT-5.4 nano     & 50 & .648 & .857 & \cellcolor[HTML]{EAF3DE}\textcolor[HTML]{3B6D11}{$+$.209} & .012 & $[+.184,\ +.233]$ & $<.001^{**}$ \\
& Qwen2.5-VL 72B   & 50 & .542 & .760 & \cellcolor[HTML]{EAF3DE}\textcolor[HTML]{3B6D11}{$+$.218} & .016 & $[+.188,\ +.250]$ & $<.001^{**}$ \\
& MiniMax-M3       & 50 & .613 & .677 & \cellcolor[HTML]{EAF3DE}\textcolor[HTML]{3B6D11}{$+$.064} & .095 & $[-.119,\ +.227]$ & $.180$ \\
& MiniMax-M2.7     & 50 & .615 & .697 & \cellcolor[HTML]{EAF3DE}\textcolor[HTML]{3B6D11}{$+$.082} & .032 & $[+.044,\ +.145]$ & $.250$ \\
& MiniMax-M2.7 HS  & 50 & .486 & .619 & \cellcolor[HTML]{EAF3DE}\textcolor[HTML]{3B6D11}{$+$.133} & .027 & $[+.081,\ +.184]$ & $<.001^{**}$ \\
\bottomrule
\end{tabular}
\caption{Model-level paired statistics (Part~II: T4--T5).
Conventions as in Table~B2.
Green = positive $\Delta$; red = degradation.
Claude Opus 4.8 / T4: $+0.012$ (95\% CI $[-.081,+.103]$, SE$=.047$); text-only
baseline $.420$ is not elevated, ruling out ceiling effects; high SE is
consistent with high scenario-level variance; cause uncertain.
$^{*}p<.05$; $^{**}p<.01$.}
\label{tab:stats_model_2}
\end{table*}

\subsection{Run-level Reproducibility}
\label{app:reproducibility}

Main results use a single evaluation run per cell.
To assess run-level stability we re-evaluated \textbf{50 scenario--model
pairs per task (250 pairs total)} across three independent runs
(temperature $=1.0$, no fixed seed).

\begin{table}[h]
\centering
\renewcommand{\arraystretch}{1.42}
\setlength{\tabcolsep}{4pt}
\small
\begin{tabular}{@{}lcccccc@{}}
\toprule
\rowcolor[HTML]{F5F4F0}
\textbf{Task} & \textbf{Run 1} & \textbf{Run 2} & \textbf{Run 3} &
\textbf{Mean} & \textbf{SD} & \textbf{95\% CI$^\dagger$} \\
\midrule
T1 Diagnose   & $+0.060$ & $+0.041$ & $+0.075$ & $+0.059$ & $0.017$ & $[+0.016,\ +0.101]$ \\
T2 Prioritize & $+0.098$ & $+0.127$ & $+0.104$ & $+0.110$ & $0.015$ & $[+0.072,\ +0.148]$ \\
T3 Allocate   & $-0.093$ & $-0.108$ & $-0.064$ & $-0.088$ & $0.022$ & $[-0.144,\ -0.033]$ \\
T4 Forecast   & $+0.231$ & $+0.236$ & $+0.225$ & $+0.231$ & $0.006$ & $[+0.217,\ +0.244]$ \\
T5 Justify    & $+0.130$ & $+0.189$ & $+0.180$ & $+0.166$ & $0.032$ & $[+0.087,\ +0.245]$ \\
\bottomrule
\end{tabular}
\caption{Run-level reproducibility (50 pairs per task, 250 total).
$^\dagger$ 95\% CI $=$ mean $\pm\,t_{0.975,2}\times\mathrm{SD}/\sqrt{3}$,
$t_{0.975,2}=4.303$; captures run-to-run variance only.
Subset means differ from full-dataset estimates due to stratified sampling.
All three T3 runs are negative ($-0.093$, $-0.108$, $-0.064$) and the
cross-run CI excludes zero.
T5 run-level variance (SD$=0.032$) is consistent with MiniMax-M3's
scenario-level SE$=.095$ in Table~B3.}
\label{tab:reproducibility}
\end{table}

A temperature-$0$ check on T3 across all 9 models on the full 50-scenario
set (450 pairs) confirms the degradation persists under deterministic
decoding (mean $\Delta=-0.081$), ruling out sampling randomness.

\section{Extended Visual Channel Ablation}
\label{app:ablation}

Section~4.5 reports ablation results for MiniMax-M3 as the primary
representative model; full cross-model results appear in
Table~C1.

The crowding pattern on T3 is consistent across all 9 models: for every
model, the best single-channel condition outperforms the full multimodal
condition (Table~C2).
Averaged across models, the best single channel improves T3 by $+0.052$
over text-only, whereas the full three-panel condition degrades it by
$-0.077$.
On T4 Forecast the inverse pattern holds for 8 of 9 models.

\begin{table*}[h]
\centering
\renewcommand{\arraystretch}{1.42}
\setlength{\tabcolsep}{5pt}
\small
\begin{tabular}{@{}llccccc@{}}
\toprule
\rowcolor[HTML]{F5F4F0}
\textbf{Task} & \textbf{Model} & \textbf{Text-only} & \textbf{Finance only} & \textbf{Growth only} & \textbf{Ops only} & \textbf{Full MM} \\
\midrule
\multicolumn{7}{@{}l}{\footnotesize\textit{T3 Resource Reallocation}} \\[1pt]
& Claude Opus 4.8  & .685 & \textbf{.755} & .728 & .742 & \cellcolor[HTML]{FCEBEB}\textcolor[HTML]{A32D2D}{.620} \\
& Gemini 3 Flash   & .635 & \textbf{.710} & .692 & .684 & \cellcolor[HTML]{FCEBEB}\textcolor[HTML]{A32D2D}{.575} \\
& Mistral Small    & .580 & .648          & .636 & \textbf{.652} & \cellcolor[HTML]{FCEBEB}\textcolor[HTML]{A32D2D}{.520} \\
& GPT-5.4 mini     & .720 & \textbf{.800} & .768 & .741 & \cellcolor[HTML]{FCEBEB}\textcolor[HTML]{A32D2D}{.645} \\
& GPT-5.4 nano     & .570 & .635          & .642 & \textbf{.649} & \cellcolor[HTML]{FCEBEB}\textcolor[HTML]{A32D2D}{.515} \\
& Qwen2.5-VL 72B   & .760 & .781          & .776 & \textbf{.789} & \cellcolor[HTML]{FCEBEB}\textcolor[HTML]{A32D2D}{.700} \\
& MiniMax-M3       & .670 & .693          & .708 & \textbf{.711} & \cellcolor[HTML]{FCEBEB}\textcolor[HTML]{A32D2D}{.542} \\
& MiniMax-M2.7     & .648 & .612          & \textbf{.655} & .641 & \cellcolor[HTML]{FCEBEB}\textcolor[HTML]{A32D2D}{.565} \\
& MiniMax-M2.7 HS  & .695 & .701          & \textbf{.714} & .709 & \cellcolor[HTML]{FCEBEB}\textcolor[HTML]{A32D2D}{.579} \\
\midrule
\multicolumn{7}{@{}l}{\footnotesize\textit{T4 Risk Forecasting}} \\[1pt]
& Claude Opus 4.8  & .420 & \textbf{.446} & .372 & .381 & \cellcolor[HTML]{EAF3DE}\textcolor[HTML]{3B6D11}{.432} \\
& Gemini 3 Flash   & .389 & \textbf{.571} & .467 & .523 & \cellcolor[HTML]{EAF3DE}\textcolor[HTML]{3B6D11}{.670} \\
& Mistral Small    & .448 & \textbf{.552} & .477 & .535 & \cellcolor[HTML]{EAF3DE}\textcolor[HTML]{3B6D11}{.716} \\
& GPT-5.4 mini     & .567 & \textbf{.704} & .647 & .631 & \cellcolor[HTML]{EAF3DE}\textcolor[HTML]{3B6D11}{.787} \\
& GPT-5.4 nano     & .568 & \textbf{.691} & .637 & .654 & \cellcolor[HTML]{EAF3DE}\textcolor[HTML]{3B6D11}{.754} \\
& Qwen2.5-VL 72B   & .408 & \textbf{.519} & .425 & .518 & \cellcolor[HTML]{EAF3DE}\textcolor[HTML]{3B6D11}{.656} \\
& MiniMax-M3       & .491 & \textbf{.623} & .589 & .613 & \cellcolor[HTML]{EAF3DE}\textcolor[HTML]{3B6D11}{.776} \\
& MiniMax-M2.7     & .392 & \textbf{.474} & .451 & .469 & \cellcolor[HTML]{EAF3DE}\textcolor[HTML]{3B6D11}{.757} \\
& MiniMax-M2.7 HS  & .372 & .438          & .421 & \textbf{.446} & \cellcolor[HTML]{EAF3DE}\textcolor[HTML]{3B6D11}{.685} \\
\bottomrule
\end{tabular}
\caption{Full cross-model visual channel ablation ($n=50$ per condition).
Bold = best single-channel per model.
Full MM: green = gain over text-only; red = loss.
The best single channel exceeds text-only for all 9 models on T3;
individual channels may fall below it (e.g.\ MiniMax-M2.7 finance-only
$.612$ vs.\ $.648$), but this pattern differs from context-length predictions.}
\label{tab:ablation_full}
\end{table*}

\begin{table*}[h]
\centering
\renewcommand{\arraystretch}{1.40}
\setlength{\tabcolsep}{5pt}
\small
\begin{tabular}{@{}llccccc@{}}
\toprule
\rowcolor[HTML]{F5F4F0}
\textbf{Task} & \textbf{Model} & \textbf{Text} & \textbf{Best Single} & \textbf{Full MM}
  & \textbf{$\Delta_{\text{full}}$} & \textbf{$\Delta_{\text{best}}$} \\
\midrule
\multicolumn{7}{@{}l}{\footnotesize\textit{T3 Allocate --- uniform crowding}} \\[1pt]
& Claude Opus 4.8  & .685 & .755 & .620 & \cellcolor[HTML]{FCEBEB}\textcolor[HTML]{A32D2D}{$-$.065} & \cellcolor[HTML]{FCEBEB}\textcolor[HTML]{A32D2D}{$-$.135} \\
& Gemini 3 Flash   & .635 & .710 & .575 & \cellcolor[HTML]{FCEBEB}\textcolor[HTML]{A32D2D}{$-$.060} & \cellcolor[HTML]{FCEBEB}\textcolor[HTML]{A32D2D}{$-$.135} \\
& Mistral Small    & .580 & .652 & .520 & \cellcolor[HTML]{FCEBEB}\textcolor[HTML]{A32D2D}{$-$.060} & \cellcolor[HTML]{FCEBEB}\textcolor[HTML]{A32D2D}{$-$.132} \\
& GPT-5.4 mini     & .720 & .800 & .645 & \cellcolor[HTML]{FCEBEB}\textcolor[HTML]{A32D2D}{$-$.075} & \cellcolor[HTML]{FCEBEB}\textcolor[HTML]{A32D2D}{$-$.155} \\
& GPT-5.4 nano     & .570 & .649 & .515 & \cellcolor[HTML]{FCEBEB}\textcolor[HTML]{A32D2D}{$-$.055} & \cellcolor[HTML]{FCEBEB}\textcolor[HTML]{A32D2D}{$-$.134} \\
& Qwen2.5-VL 72B   & .760 & .789 & .700 & \cellcolor[HTML]{FCEBEB}\textcolor[HTML]{A32D2D}{$-$.060} & \cellcolor[HTML]{FCEBEB}\textcolor[HTML]{A32D2D}{$-$.089} \\
& MiniMax-M3       & .670 & .711 & .542 & \cellcolor[HTML]{FCEBEB}\textcolor[HTML]{A32D2D}{$-$.128} & \cellcolor[HTML]{FCEBEB}\textcolor[HTML]{A32D2D}{$-$.169} \\
& MiniMax-M2.7     & .648 & .655 & .565 & \cellcolor[HTML]{FCEBEB}\textcolor[HTML]{A32D2D}{$-$.083} & \cellcolor[HTML]{FCEBEB}\textcolor[HTML]{A32D2D}{$-$.090} \\
& MiniMax-M2.7 HS  & .695 & .714 & .579 & \cellcolor[HTML]{FCEBEB}\textcolor[HTML]{A32D2D}{$-$.106} & \cellcolor[HTML]{FCEBEB}\textcolor[HTML]{A32D2D}{$-$.135} \\
\midrule
\multicolumn{7}{@{}l}{\footnotesize\textit{T4 Forecast --- cross-channel complementarity}} \\[1pt]
& Claude Opus 4.8  & .420 & .446 & .432 & \cellcolor[HTML]{EAF3DE}\textcolor[HTML]{3B6D11}{$+$.012} & \cellcolor[HTML]{FCEBEB}\textcolor[HTML]{A32D2D}{$-$.014} \\
& Gemini 3 Flash   & .389 & .571 & .670 & \cellcolor[HTML]{EAF3DE}\textcolor[HTML]{3B6D11}{$+$.281} & \cellcolor[HTML]{EAF3DE}\textcolor[HTML]{3B6D11}{$+$.099} \\
& Mistral Small    & .448 & .552 & .716 & \cellcolor[HTML]{EAF3DE}\textcolor[HTML]{3B6D11}{$+$.268} & \cellcolor[HTML]{EAF3DE}\textcolor[HTML]{3B6D11}{$+$.164} \\
& GPT-5.4 mini     & .567 & .704 & .787 & \cellcolor[HTML]{EAF3DE}\textcolor[HTML]{3B6D11}{$+$.220} & \cellcolor[HTML]{EAF3DE}\textcolor[HTML]{3B6D11}{$+$.083} \\
& GPT-5.4 nano     & .568 & .691 & .754 & \cellcolor[HTML]{EAF3DE}\textcolor[HTML]{3B6D11}{$+$.186} & \cellcolor[HTML]{EAF3DE}\textcolor[HTML]{3B6D11}{$+$.063} \\
& Qwen2.5-VL 72B   & .408 & .519 & .656 & \cellcolor[HTML]{EAF3DE}\textcolor[HTML]{3B6D11}{$+$.248} & \cellcolor[HTML]{EAF3DE}\textcolor[HTML]{3B6D11}{$+$.137} \\
& MiniMax-M3       & .491 & .623 & .776 & \cellcolor[HTML]{EAF3DE}\textcolor[HTML]{3B6D11}{$+$.285} & \cellcolor[HTML]{EAF3DE}\textcolor[HTML]{3B6D11}{$+$.153} \\
& MiniMax-M2.7     & .392 & .474 & .757 & \cellcolor[HTML]{EAF3DE}\textcolor[HTML]{3B6D11}{$+$.365} & \cellcolor[HTML]{EAF3DE}\textcolor[HTML]{3B6D11}{$+$.283} \\
& MiniMax-M2.7 HS  & .372 & .446 & .685 & \cellcolor[HTML]{EAF3DE}\textcolor[HTML]{3B6D11}{$+$.313} & \cellcolor[HTML]{EAF3DE}\textcolor[HTML]{3B6D11}{$+$.239} \\
\bottomrule
\end{tabular}
\caption{Cross-model crowding scope summary from Table~C1.
$\Delta_{\text{full}}=\text{Full MM}-\text{Text}$;
$\Delta_{\text{best}}=\text{Full MM}-\text{Best Single}$.}
\label{tab:crowding_full}
\end{table*}

\section{Alternative Hypotheses: Context Length and Numerical Density}
\label{app:confound}

\subsection{Context Length as Confound}

One alternative explanation for the T3 degradation is that adding visual
inputs increases total input length.
We address this through two observations.
First, the best single-channel condition outperforms text-only for all 9
models on T3 (Table~C2), even though each single
channel also adds a visual image.
Second, T4 Forecast receives the same full multimodal input as T3 yet shows
substantial positive uplift.
The task-specificity is inconsistent with a context-length explanation.

\subsection{Text-plus-Numbers Control}
\label{app:text_numbers}

To test whether the T3 effect is driven by numerical density rather than
visual modality, we ran a supplementary control in which the key numerical
values from each panel were appended as structured text, without charts.
Table~D1 decomposes each task's total uplift into
$\Delta_{\text{num}}$ (text-plus-numbers vs.\ text-only) and
$\Delta_{\text{vis}}=\Delta_{\text{full MM}}-\Delta_{\text{num}}$.

\begin{table}[h]
\centering
\renewcommand{\arraystretch}{1.42}
\small
\begin{tabular}{@{}lccc@{}}
\toprule
\rowcolor[HTML]{F5F4F0}
\textbf{Task} & \textbf{$\Delta_{\text{num}}$} &
\textbf{$\Delta_{\text{vis}}$} & \textbf{$\Delta_{\text{full MM}}$} \\
\midrule
T1 Diagnose   & $+0.031$ & $+0.038$ & $+0.069$ \\
T2 Prioritize & $+0.044$ & $+0.054$ & $+0.098$ \\
T3 Allocate   & $-0.046$ & $-0.031$ & $-0.077$ \\
T4 Forecast   & $+0.102$ & $+0.136$ & $+0.237$ \\
T5 Justify    & $+0.071$ & $+0.099$ & $+0.170$ \\
\bottomrule
\end{tabular}
\caption{Task-level decomposition of multimodal uplift into numerical
($\Delta_{\text{num}}$) and visual ($\Delta_{\text{vis}}$) components,
averaged across all 9 models.
Per-model decompositions with sign tests for T3 and T4 in
Tables~D2--D3.}
\label{tab:text_numbers_task}
\end{table}

For T4 and T5, the visual component accounts for the majority of uplift
($\Delta_{\text{vis}}=+0.136$ and $+0.099$), confirming that these gains
depend on visual trend representations rather than underlying numbers alone.
For T3, 60\% of the degradation comes from the numerical component
($\Delta_{\text{num}}=-0.046$), making the T3 failure primarily an
\textbf{evidence-load effect} with visual presentation acting as an amplifier.
We define \textbf{visual signal crowding} as the specific instantiation of
evidence-load crowding in which the amplifying signals are visual charts.

\begin{table*}[h]
\centering
\renewcommand{\arraystretch}{1.38}
\setlength{\tabcolsep}{5pt}
\small
\begin{tabular}{@{}lcccc@{}}
\toprule
\rowcolor[HTML]{F5F4F0}
\textbf{Model} & \textbf{$\Delta_{\text{full}}$} &
\textbf{$\Delta_{\text{num}}$} & \textbf{$\Delta_{\text{vis}}$} & \textbf{Num\%} \\
\midrule
Claude Opus 4.8  & $-0.065$ & $-0.038$ & $-0.027$ & 58\% \\
Gemini 3 Flash   & $-0.060$ & $-0.031$ & $-0.029$ & 52\% \\
Mistral Small    & $-0.060$ & $-0.039$ & $-0.021$ & 65\% \\
GPT-5.4 mini     & $-0.075$ & $-0.046$ & $-0.029$ & 61\% \\
GPT-5.4 nano     & $-0.055$ & $-0.040$ & $-0.015$ & 73\% \\
Qwen2.5-VL 72B   & $-0.060$ & $-0.039$ & $-0.021$ & 65\% \\
MiniMax-M3       & $-0.125$ & $-0.069$ & $-0.056$ & 55\% \\
MiniMax-M2.7     & $-0.083$ & $-0.041$ & $-0.042$ & 49\% \\
MiniMax-M2.7 HS  & $-0.106$ & $-0.071$ & $-0.035$ & 67\% \\
\midrule
\rowcolor[HTML]{F5F4F0}
\textbf{Mean}    & $-0.077$ & $-0.046$ & $-0.031$ & \textbf{60\%} \\
\bottomrule
\end{tabular}
\caption{Per-model T3 decomposition ($n=50$ each).
Num\% = ratio of means ($-0.046/-0.077=60\%$; per-model range 49\%--73\%).
$\Delta_{\text{vis}}<0$ for all 9 models (sign test: $k=9/9$, $p=0.004$).
Per-model decomposition reported for T3 and T4 only, as they constitute
the core sign-reversal contrast; T1, T2, T5 task-level decompositions in
Table~D1.}
\label{tab:text_numbers_model_t3}
\end{table*}

\begin{table*}[h]
\centering
\renewcommand{\arraystretch}{1.38}
\setlength{\tabcolsep}{5pt}
\small
\begin{tabular}{@{}lcccc@{}}
\toprule
\rowcolor[HTML]{F5F4F0}
\textbf{Model} & \textbf{$\Delta_{\text{full}}$} &
\textbf{$\Delta_{\text{num}}$} & \textbf{$\Delta_{\text{vis}}$} & \textbf{Vis\%} \\
\midrule
Claude Opus 4.8$^\dagger$ & $+0.012$ & $+0.005$ & $+0.007$ & --- \\
Gemini 3 Flash   & $+0.281$ & $+0.120$ & $+0.161$ & 57\% \\
Mistral Small    & $+0.268$ & $+0.094$ & $+0.174$ & 65\% \\
GPT-5.4 mini     & $+0.220$ & $+0.091$ & $+0.129$ & 59\% \\
GPT-5.4 nano     & $+0.186$ & $+0.071$ & $+0.115$ & 62\% \\
Qwen2.5-VL 72B   & $+0.248$ & $+0.110$ & $+0.138$ & 56\% \\
MiniMax-M3       & $+0.286$ & $+0.132$ & $+0.154$ & 54\% \\
MiniMax-M2.7     & $+0.323$ & $+0.127$ & $+0.196$ & 61\% \\
MiniMax-M2.7 HS  & $+0.313$ & $+0.165$ & $+0.148$ & 47\% \\
\midrule
\rowcolor[HTML]{F5F4F0}
\textbf{Mean (excl.\ Claude)} & $+0.266$ & $+0.114$ & $+0.152$ & \textbf{57\%} \\
\bottomrule
\end{tabular}
\caption{Per-model T4 decomposition ($n=50$ each).
Vis\% = ratio of means, excluding Claude.
$^\dagger$ Claude's $\Delta_{\text{full}}=+0.012$ (CI $[-0.081,+0.103]$)
is indistinguishable from zero; Vis\% not reported; even excluding Claude,
$\Delta_{\text{vis}}>0$ for all 8 remaining models ($k=8/8$, $p=0.008$).
Including Claude the count is $9/9$; we report $8/8$ as the conservative
estimate.
The nine-model mean $\Delta_{\text{vis}}=+0.136$ matches
Table~D1.
Symmetric pattern: $\Delta_{\text{vis}}<0$ for all 9 on T3 ($p=0.004$);
$\Delta_{\text{vis}}>0$ for 8/8 non-trivial models on T4 ($p=0.008$).}
\label{tab:text_numbers_model_t4}
\end{table*}

\subsection{Toward a Complete Load $\times$ Modality Design}
\label{app:2x2}

Three cells of a $2\times2$ design (load $\times$ modality) have been
measured on T3:

\begin{itemize}[leftmargin=1.4em, itemsep=2pt, topsep=3pt]
  \item \textbf{1 channel, visual} (best single-channel): $\Delta=+0.052$; positive for all 9 models.
  \item \textbf{3 channels, text} (text-plus-numbers): $\Delta=-0.046$.
  \item \textbf{3 channels, visual} (full multimodal): $\Delta=-0.077$; negative for all 9 models.
\end{itemize}

The three cells are consistent with the evidence-load account: increasing
concurrent channels reverses the sign under both modalities relative to a
single visual channel, and at equal load the visual condition is $0.031$
worse than the textual one.
The fourth cell (single-panel-as-text) has not been measured; directly
running it would allow a formal load $\times$ modality interaction test and
is the primary next experiment.
Until then, our claims are limited to the ordering of the three measured
cells and the modality contrast at fixed three-channel load.

\section{Pooled Paired and Mixed-Effects Analysis for T3}
\label{app:t3_pooled}

Across \textbf{444} valid paired scenario--model observations (sum of
per-model $n$ in Table~B2; 6 of 450 nominal excluded
due to missing or unparsable outputs), T3 exhibits a statistically reliable
negative multimodal effect.

\begin{table}[h]
\centering
\renewcommand{\arraystretch}{1.4}
\small
\begin{tabular}{@{}lccc@{}}
\toprule
\rowcolor[HTML]{F5F4F0}
\textbf{Analysis} & \textbf{Estimate} & \textbf{95\% CI / SE} & \textbf{$p$} \\
\midrule
Model-level mean $\Delta$     & $-0.077$ & $[-0.093,\ -0.063]$ & $0.004^{**}$ \\
Pooled paired $\Delta$        & $-0.078$ & $[-0.090,\ -0.066]$ & $<0.001^{**}$ \\
Mixed-effects modality effect & $-0.074$ & $\mathrm{SE}=0.018$ & $<0.001^{**}$ \\
\bottomrule
\end{tabular}
\caption{Robustness checks for T3 ($n=444$ valid pairs).
Model-level CI ($[-0.093,-0.063]$, bootstrapped over $n=9$ means) is wider
than pooled CI ($[-0.090,-0.066]$, bootstrapped over $n=444$ pairs),
indicating greater between-model than within-model heterogeneity.
The $0.001$ difference in point estimates arises from model-count vs.\
observation-count weighting with near-balanced group sizes ($n=46$--$50$).
$^{**}p<0.01$.}
\label{tab:t3_pooled_target}
\end{table}

\begin{table}[h]
\centering
\renewcommand{\arraystretch}{1.4}
\small
\begin{tabular}{@{}lccc@{}}
\toprule
\rowcolor[HTML]{F5F4F0}
\textbf{T3 Family} & \textbf{Mean $\Delta$} & \textbf{95\% CI} & \textbf{$n_{\text{paired}}$} \\
\midrule
Overall      & $-0.078$ & $[-0.090,\ -0.066]$ & 444 \\
Easy         & $-0.028$ & $[-0.054,\ -0.006]$ & 112 \\
Fragile      & $-0.061$ & $[-0.092,\ -0.028]$ & 111 \\
Tension      & $-0.118$ & $[-0.154,\ -0.081]$ & 112 \\
Adversarial  & $-0.102$ & $[-0.141,\ -0.064]$ & 109 \\
\bottomrule
\end{tabular}
\caption{Pooled family-level results for T3
($112+111+112+109=444$).
Strongest in tension and adversarial settings; directionally consistent
across all families.
The Overall row is the direct pooled bootstrap estimate; the $n$-weighted
combination of the four displayed (three-decimal) family values gives
$-0.0771$, agreeing with the Overall row to two decimal places.}
\label{tab:t3_family_breakdown_target}
\end{table}

\section{Cross-Model Scope of the Crowding Effect}
\label{app:crowding_scope}

For all 9 models: full multimodal falls below the best single channel on T3,
and falls below text-only.
For T4: full multimodal outperforms every single channel for 8 of 9 models.

\begin{table}[h]
\centering
\renewcommand{\arraystretch}{1.35}
\small
\begin{tabular}{@{}lccc@{}}
\toprule
\rowcolor[HTML]{F5F4F0}
\textbf{Task} & \textbf{Criterion} & \textbf{Model Support} & \textbf{Interpretation} \\
\midrule
T3 Allocate & Full MM $<$ best single-channel & 9/9 & Uniform crowding \\
T3 Allocate & Full MM $<$ text-only           & 9/9 & Uniform degradation \\
T4 Forecast & Full MM $>$ all single-channel  & 8/9 & Cross-channel complementarity \\
\bottomrule
\end{tabular}
\caption{Cross-model scope of the channel interaction effect.}
\label{tab:crowding_scope_target}
\end{table}

\section{Additional Behavioral Evidence}
\label{app:crowding_behavioral}

\paragraph{Sub-dimension interaction within T3.}
Full multimodal continues to improve \textit{visual\_grounding\_proxy} but
deteriorates \textit{validity}, \textit{reallocation\_accuracy}, and
\textit{strategic\_fit} (Figure~G2), consistent with
coordination failure rather than perception failure.

\paragraph{Difficulty-conditioned interaction.}
The T3 multimodal effect is negative across all four families
(Table~E2) and largest under tension and
adversarial settings, suggesting crowding intensifies when competing signals
must be reconciled under explicit trade-offs.

\paragraph{Constraint-failure concentration.}
At a qualitative level, the T3 drop appears concentrated in
constraint-relevant failures while visual grounding remains strong.

\paragraph{Evidence-load analysis.}
Full multimodal outputs cite more evidence than the best single channel
in both tasks (T3: $2.6\to4.1$; T4: $2.8\to4.0$; Table~G1).
In T3 this co-occurs with lower validity ($.89\to.72$) and strategic fit
($.81\to.64$); in T4 with higher leading-indicator ($.60\to.75$) and risk
recall ($.63\to.78$).

\begin{table*}[h]
\centering
\renewcommand{\arraystretch}{1.35}
\setlength{\tabcolsep}{5pt}
\small
\begin{tabular}{@{}llccc@{}}
\toprule
\rowcolor[HTML]{F5F4F0}
\textbf{Task} & \textbf{Condition} & \textbf{Cited evidence} & \textbf{Validity / LI recall} & \textbf{Strategic fit / Risk recall} \\
\midrule
T3 Allocate & Best Single & 2.6 & .89 & .81 \\
T3 Allocate & Full MM     & 4.1 & .72 & .64 \\
\midrule
T4 Forecast & Best Single & 2.8 & .60 & .63 \\
T4 Forecast & Full MM     & 4.0 & .75 & .78 \\
\bottomrule
\end{tabular}
\caption{Evidence-load summary. T3: validity/strategic fit columns.
T4: leading-indicator/risk recall columns.}
\label{tab:evidence_load}
\end{table*}

\paragraph{Leave-one-out panel-drop analysis.}
Removing any single panel from full multimodal partially restores T3 quality
(Table~G2); removing panels from T4 full multimodal lowers
it.

\begin{table*}[h]
\centering
\renewcommand{\arraystretch}{1.32}
\setlength{\tabcolsep}{5pt}
\small
\begin{tabular}{@{}llcccc@{}}
\toprule
\rowcolor[HTML]{F5F4F0}
\textbf{Task} & \textbf{Model / Condition} & \textbf{Score} & \textbf{Validity / LI} & \textbf{Strat.\ fit / Risk} & \textbf{Interpretation} \\
\midrule
T3 & MiniMax-M3 Full MM      & .542 & .64 & .58 & Baseline crowding \\
   & MiniMax-M3 $-$ Finance  & .624 & .76 & .69 & Partial recovery \\
   & MiniMax-M3 $-$ Growth   & .603 & .73 & .66 & Partial recovery \\
   & MiniMax-M3 $-$ Ops      & .579 & .69 & .62 & Smallest recovery \\
\cmidrule(lr){2-6}
   & GPT-5.4 mini Full MM    & .645 & .71 & .63 & Baseline crowding \\
   & GPT-5.4 mini $-$ Finance& .724 & .82 & .73 & Partial recovery \\
   & GPT-5.4 mini $-$ Growth & .701 & .79 & .70 & Partial recovery \\
   & GPT-5.4 mini $-$ Ops    & .676 & .75 & .66 & Smallest recovery \\
\midrule
T4 & MiniMax-M3 Full MM      & .776 & .76 & .78 & Full gain \\
   & MiniMax-M3 $-$ Finance  & .724 & .70 & .72 & Drop from full MM \\
   & MiniMax-M3 $-$ Growth   & .706 & .68 & .70 & Drop from full MM \\
   & MiniMax-M3 $-$ Ops      & .714 & .69 & .71 & Drop from full MM \\
\cmidrule(lr){2-6}
   & GPT-5.4 mini Full MM    & .787 & .77 & .75 & Full gain \\
   & GPT-5.4 mini $-$ Finance& .741 & .72 & .69 & Drop from full MM \\
   & GPT-5.4 mini $-$ Growth & .723 & .69 & .67 & Drop from full MM \\
   & GPT-5.4 mini $-$ Ops    & .731 & .70 & .68 & Drop from full MM \\
\bottomrule
\end{tabular}
\caption{Leave-one-out panel-drop analysis.
T3 metric columns: validity / strategic fit.
T4 metric columns: leading-indicator / risk recall.}
\label{tab:panel_drop}
\end{table*}

\begin{figure*}[h]
  \centering
  \includegraphics[width=\linewidth]{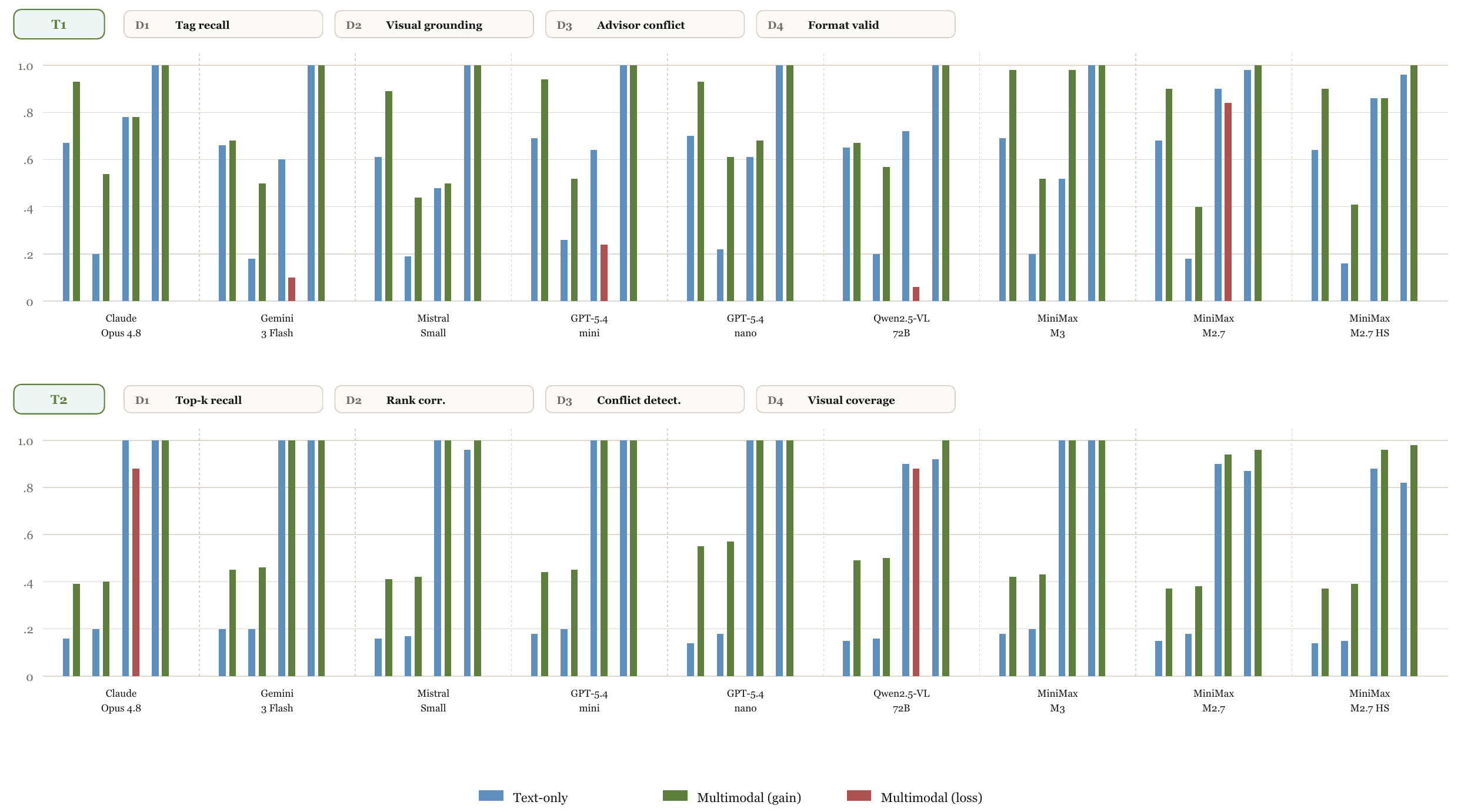}
  \caption{Per-model sub-dimension scores for T1 and T2.}
  \label{fig:app_subdim_T1T2}
\end{figure*}

\begin{figure*}[h]
  \centering
  \includegraphics[width=\linewidth]{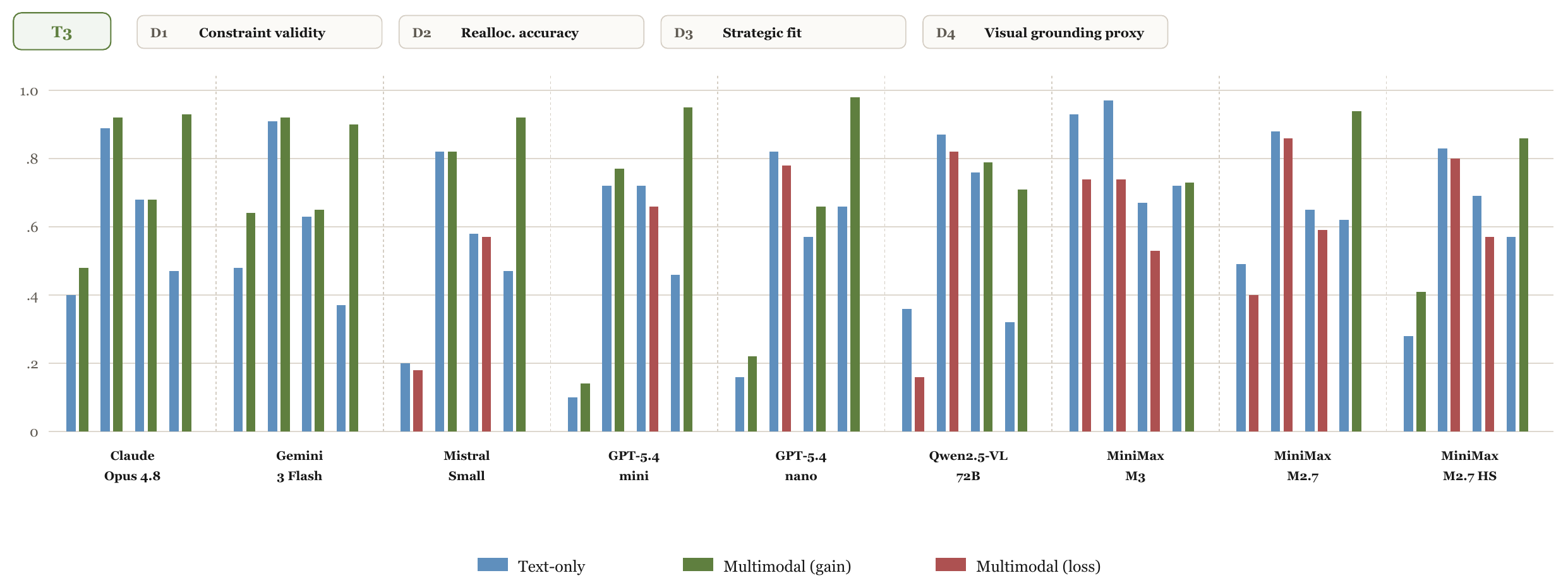}
  \caption{Per-model sub-dimension scores for T3.
  Visual grounding proxy rises while constraint validity and strategic fit
  decline across models.}
  \label{fig:app_subdim_T3}
\end{figure*}

\begin{figure*}[h]
  \centering
  \includegraphics[width=\linewidth]{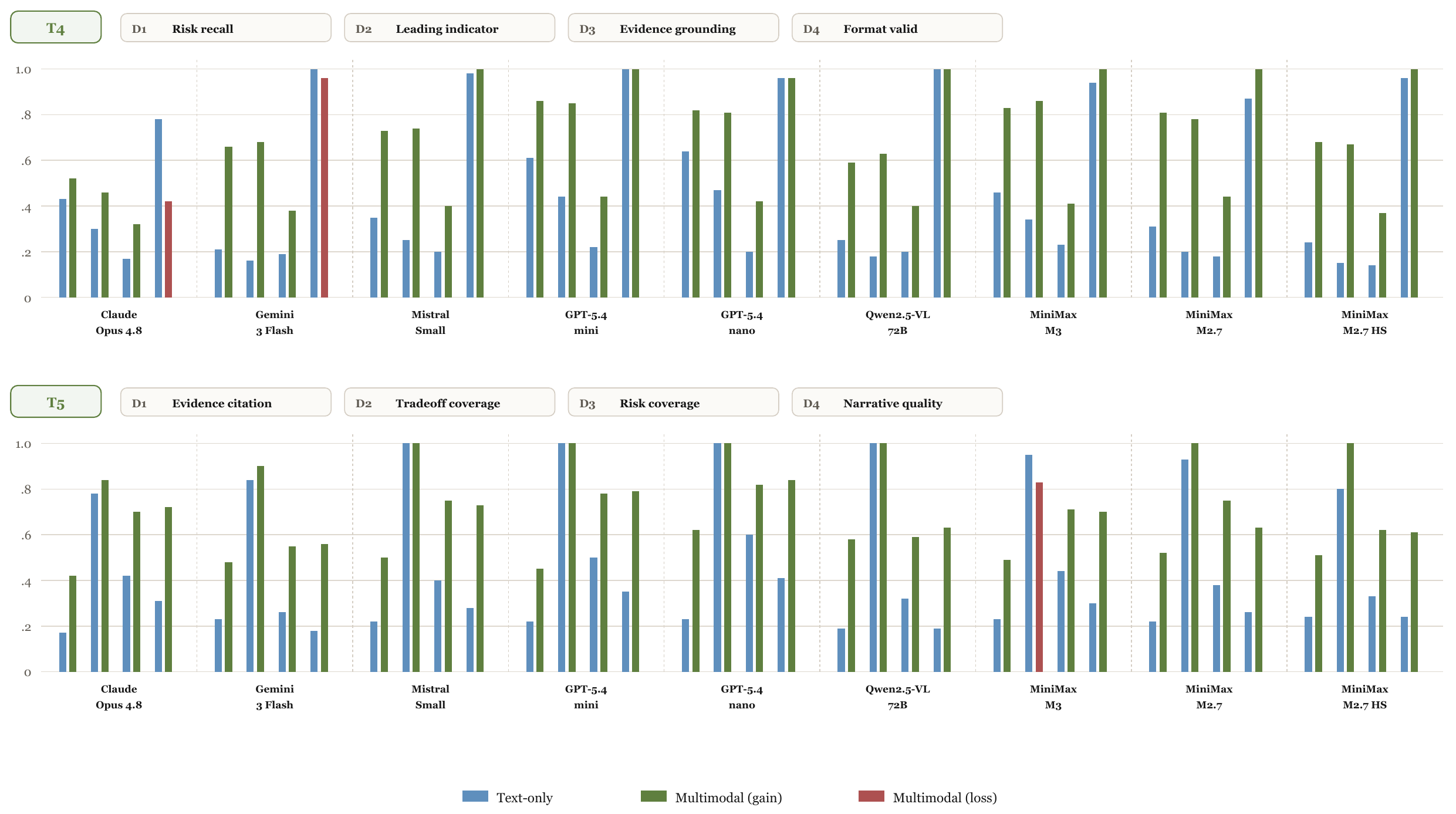}
  \caption{Per-model sub-dimension scores for T4 and T5.}
  \label{fig:app_subdim_T4T5}
\end{figure*}

\section{Discussion}
\label{app:discussion}

\subsection{Implications for Multimodal Evaluation}

Our results suggest that current multimodal evaluation practices conflate
two fundamentally different capabilities: perceiving visual information and
acting upon it (Figure~H1).
Our findings suggest a three-stage capability hierarchy: (1) perceiving
visual evidence, (2) integrating it into coherent reasoning, and
(3) transforming reasoning into valid constrained actions.
Future benchmarks should distinguish these as separate evaluation dimensions.

\subsection{Evidence-Load Crowding Beyond Executive Decision Making}

\textbf{Evidence-load crowding} describes the failure of constrained action
arising when multiple concurrent evidence signals compete during execution,
regardless of modality.
\textbf{Visual signal crowding} is the specific instantiation where the
amplifying signals are visual charts.
The text-plus-numbers control locates the mechanism: 60\% of T3 degradation
arises from numerical density alone ($\Delta_{\text{num}}=-0.046$), with
visual presentation adding $\Delta_{\text{vis}}=-0.031$ ($p=0.004$ across
all 9 models).
On T4, $\Delta_{\text{vis}}=+0.152$ (8/8 non-trivial models, $p=0.008$;
nine-model mean $+0.136$), confirming the amplification is bidirectional.

\begin{figure*}[t]
  \centering
  \includegraphics[width=\linewidth]{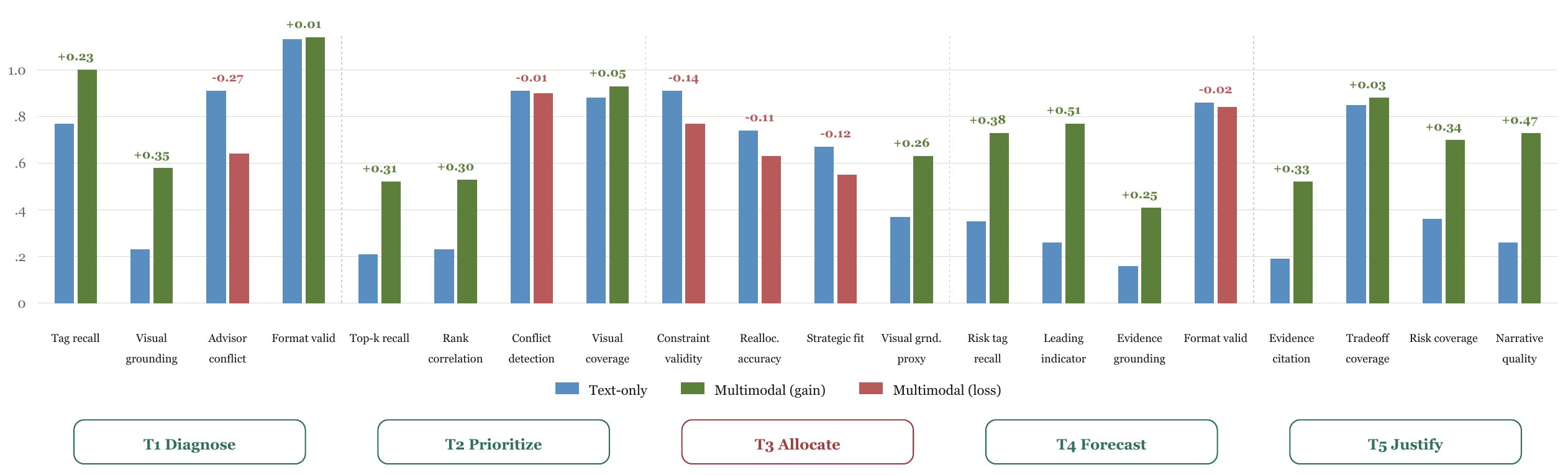}
  \caption{Evaluation sub-dimension scores under text-only and multimodal
  conditions, averaged across all 9 models.}
  \label{fig:subdimension}
\end{figure*}

\subsection{Evidence Integration as the Next Challenge}

The critical capability is not accumulating more evidence but selectively
integrating the most relevant evidence while suppressing distractors.
Progress should be measured by advances in selective evidence integration,
not only visual understanding.

\subsection{Design Principles for Future Multimodal Agents}

First, \textbf{two-stage reasoning architectures}: separate evidence
acquisition from decision execution.
Second, \textbf{selective evidence exposure}: the text-plus-numbers control
shows 60\% of the T3 degradation survives removal of the visual modality,
so evidence selection applies to quantitative signals generally.
Third, \textbf{constraint-aware decision execution}: preserve task
constraints independently of the surrounding context.

\section{Limitations and Future Work}
\label{app:limitations}

\subsection{Ceiling Effects and Headroom Asymmetry}

Proportional uplift $\Delta/(1-\text{baseline})$ for positive tasks:
T1 $0.205$, T2 $0.330$, T4 $0.437$, T5 $0.388$.
T4 remains strongest after normalisation.
For T3, proportional degradation relative to baseline:
$-0.077/0.661=-0.117$; not compared directly to positive-direction figures
due to different denominators.

\subsection{Model Pool Composition}

T3 degradation present in all 9 models ($-0.055$ to $-0.125$), with no
monotonic relationship to model scale.
Future work should include reasoning-mode variants (e.g.\ o-series,
extended-thinking modes).

\subsection{Scope and Remaining Controls}

The load $\times$ modality design is incomplete (Appendix~D.3).
Ground-truth validation covers 15 of 50 T3 scenarios by authors
(Appendix~A.2).
Human--automatic agreement measured on GPT-5.4 mini only.
The multimodal condition supplies more evidence than text-only; the
text-plus-numbers control (Appendix~D.2) is the partial
remedy.

\subsection{Future Directions}

Mechanistic analyses (attention tracing, representation probing) could
complement the behavioral evidence.
Extending to clinical planning, robotics, and scientific discovery would
test generality of evidence-load crowding.

\section{Extended Related Work}
\label{app:related}

\subsection{LLMs as Decision-Making Agents}

Chain-of-thought~\citep{wei2022chain,zhang2022automatic},
ReAct~\citep{yao2022react}, Tree of Thoughts~\citep{yao2023tree},
AgentBench~\citep{liu2024agentbench}, WebShop~\citep{yao2022webshop},
ALFWorld~\citep{shridhar2020alfworld},
$\tau$-bench~\citep{yao2024tau},
DSGBench~\citep{tang2025dsgbench}.

\begin{figure*}[t]
  \centering
  \includegraphics[width=\linewidth]{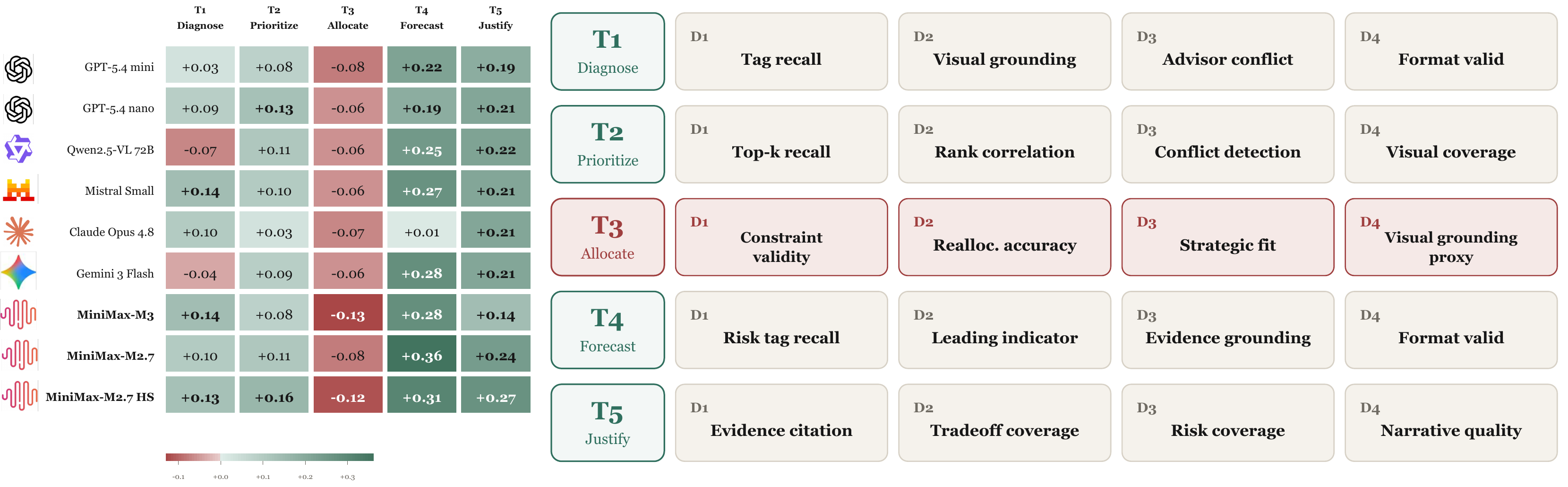}
  \caption{C-SuiteBench benchmark overview.
  \textbf{\textit{Left}} Multimodal uplift $\Delta$ across task--difficulty
  combinations; red border marks T3.
  \textbf{\textit{Right}} Per-task evaluation rubric decomposition.}
  \label{fig:grand_overview}
\end{figure*}

\subsection{Executive Decision-Making Benchmarks}

CEO-Bench~\citep{chen2026ceo}, RetailBench~\citep{zhang2026retailbench},
\citet{dai2026can}.
All existing executive benchmarks are text-only.

\subsection{Multimodal Understanding of Business Visuals}

ChartQA~\citep{masry2022chartqa}, FigureQA~\citep{kahou2017figureqa},
PlotQA~\citep{methani2020plotqa}, DocVQA~\citep{mathew2021docvqa},
OCRBench~\citep{liu2024ocrbench},
ChartInstruct~\citep{masry2024chartinstruct},
MMMU~\citep{yue2024mmmu}, MathVista~\citep{lu2024mathvista},
MMT-Bench~\citep{ying2024mmt}.
All treat visual understanding as an end in itself.

\subsection{Distraction and Irrelevant Context}

\citet{shi2023large} show LLMs are misled by distracting information.
Our text-plus-numbers control refines this: the signals that produce
crowding here are \emph{not} irrelevant --- every individual channel is
beneficial in isolation --- yet their combination disrupts constrained
allocation.

\subsection{Financial and Business Reasoning Benchmarks}

FinQA~\citep{chen2021finqa}, ConvFinQA~\citep{chen2022convfinqa},
FinanceBench~\citep{islam2023financebench},
BizBench~\citep{koncel2023bizbench},
BlueFin~\citep{kundurthy2026bluefin},
Finance Agent Benchmark~\citep{bigeard2025finance},
BizFinBench~\citep{lu2025bizfinbench},
FinTradeBench~\citep{agrawal2026fintradebench},
FIRE~\citep{zhang2026fire},
RealFin~\citep{dai2026realfin}.
All text-only; none provides paired multimodal evaluation.

\end{document}